\documentclass[11pt,a4paper]{article}
\usepackage{times,latexsym}
\usepackage{url}
\usepackage[T1]{fontenc}
\usepackage[mode=build]{standalone}
\usepackage{tikz,pgfplots,pgfplotstable}
\usepackage{soul}
\usetikzlibrary{positioning,shapes,shadows,arrows,automata,chains,fit}
\usepackage[acceptedWithA]{tacl2018v2}

\usepackage[]{tacl2018v2}

\usepackage{pslatex}
\usepackage{latexsym}
\usepackage{amsmath}
\usepackage{graphicx}  %
\graphicspath{ {./images/} }
\usepackage{multirow}
\usepackage{listings}

\usepackage{paralist}
\usepackage[inline]{enumitem}
\usepackage{amssymb}
\DeclareMathOperator\avg{\operatorname{avg}}
\DeclareMathOperator\sigmoid{\operatorname{sigmoid}}

\newcommand{\lform}[1]{\textsf{\scriptsize{#1}}}
\newcommand{\cmtt}[1]{{\fontfamily{cmtt}\selectfont {#1}}}
\newcommand{\xml}[1]{$<${\fontfamily{cmtt}\selectfont {#1}}$>$}

\DeclareMathOperator*{\argmax}{arg\,max}
\makeatletter
\newcommand{\thickhline}{%
    \noalign {\ifnum 0=`}\fi \hrule height 1pt
    \futurelet \reserved@a \@xhline
}
\makeatother
\usepackage{microtype}

\title{Data-to-text Generation with Macro Planning}

\author{Ratish Puduppully  \textnormal{and} Mirella Lapata\\
Institute for Language, Cognition and Computation\\
School of Informatics, University of Edinburgh\\
 10 Crichton Street, Edinburgh EH8 9AB\\
\texttt{r.puduppully@sms.ed.ac.uk}~~~~\texttt{mlap@inf.ed.ac.uk}\\
}

\date{}

\begin{document}
\maketitle
\begin{abstract}
  Recent approaches to data-to-text generation have adopted the very
  successful encoder-decoder architecture or variants thereof.  These
  models generate text which is fluent (but often imprecise) and
  perform quite poorly at selecting appropriate content and ordering
  it coherently.  To overcome some of these issues, we propose a
  neural model with a \emph{macro planning} stage followed by a
  generation %
  stage 
  reminiscent of traditional methods which embrace
  separate modules for planning and surface realization. 
  Macro plans
  represent high level organization of important content such as
  entities, events and their interactions; 
 they are learnt
  from data
  and given as input to the generator.
  Extensive experiments on two data-to-text benchmarks
  (\textsc{RotoWire} and MLB) show that our approach outperforms
  competitive baselines in terms of automatic and human evaluation.
\end{abstract}

\section{Introduction}

Data-to-text generation refers to the task of generating textual
output from non-linguistic input
\cite{Reiter:1997:BAN:974487.974490,reiter-dale:00,
  DBLP:journals/jair/GattK18} such as databases of records,
simulations of physical systems, accounting spreadsheets, or expert
system knowledge bases. As an example, Figure~\ref{fig:example} shows
various statistics describing a major league baseball (MLB) game,
including extracts from the \emph{box score} (i.e.,~the performance of
the two teams and individual team members who played as batters,
pitchers or fielders; Tables~(A)), \emph{play-by-play} (i.e.,~the
detailed sequence of each play of the game as it occurred; Table~(B)),
and a human written game \emph{summary} (Table~(C)).

\begin{figure*}[t!]
\begin{minipage}{0.4\textwidth}
\footnotesize
\hspace*{.1cm}\begin{tabular}{@{}l@{~~}r@{~~}c@{~~}r@{~~}c@{~~}c@{~~}c@{~~}c@{~~}c@{~~}l@{}} 
\multicolumn{10}{@{}l@{}}{(A)} \\ 
\hline
\lform{TEAM}      & \lform{Inn1} &\lform{Inn2} &\lform{Inn3} & \lform{Inn4} & \lform{$\dots$} & \lform{TR}& \lform{TH} & \lform{E} & \lform{$\dots$} \\ \hline
\lform{Orioles} &\lform{1} &\lform{0} &\lform{0} & \lform{0} & \lform{$\dots$} & \lform{2} & \lform{4} & \lform{0} & \lform{$\dots$} \\ 
\lform{Royals} &\lform{1} &\lform{0} &\lform{0} & \lform{3} &\lform{$\dots$} & \lform{9} & \lform{14} & \lform{1} & \lform{$\dots$} \\ \hline
\end{tabular}

\vspace{.3cm}
\footnotesize
\hspace*{.3cm}\begin{tabular}{@{}l@{~~}c@{~~}c@{~}r@{~~}c@{~~}c@{~~}l@{~~}l@{}} 
\hline
\lform{BATTER} & \lform{H/V} & \lform{AB} & \lform{BR} & \lform{BH}& \lform{RBI} &  \lform{TEAM} & \lform{$\dots$}\\ \hline
\lform{C.Mullins} & \lform{H} & \lform{4} & \lform{2} & \lform{2}& \lform{1} &   \lform{Orioles} & \lform{$\dots$}\\
\lform{J.Villar} & \lform{H} & \lform{4} & \lform{0} & \lform{0}& \lform{0} & \lform{Orioles} & \lform{$\dots$}\\
\lform{W.Merrifield} & \lform{V}& \lform{2} & \lform{3} & \lform{2} & \lform{1}& \lform{Royals} & \lform{$\dots$}\\
\lform{R.O'Hearn} & \lform{V}& \lform{5} & \lform{1} & \lform{3} & \lform{4}&  \lform{Royals} & \lform{$\dots$}\\

\lform{$\dots$} & \lform{$\dots$} & \lform{$\dots$} & \lform{$\dots$} &  \lform{$\dots$} &  \lform{$\dots$} & \lform{$\dots$}\\\hline
\end{tabular}

\vspace{.3cm}
\footnotesize
\begin{tabular}{@{}l@{~~}c@{~~}c@{~}r@{~~}c@{~~}c@{~~}c@{~~}c@{~~}c@{~~}l@{~~}l@{}} \hline
\lform{PITCHER} & \lform{H/V} & \lform{W} & \lform{L} & \lform{IP}& \lform{PH} &  \lform{PR} &  \lform{ER}&  \lform{BB}&  \lform{K} & \lform{$\dots$}\\ \hline
\lform{A.Cashner} & \lform{H} & \lform{4} & \lform{13} & \lform{5.1}& \lform{9} &   \lform{4}&   \lform{4} &   \lform{3}&   \lform{1}& \lform{$\dots$}\\
\lform{B.Keller} & \lform{V}& \lform{7} & \lform{5} & \lform{8.0} & \lform{4}&  \lform{2}  &   \lform{2}&   \lform{2}&   \lform{4} & \lform{$\dots$}\\
\lform{$\dots$} & \lform{$\dots$} & \lform{$\dots$} & \lform{$\dots$} &  \lform{$\dots$} &  \lform{$\dots$} & \lform{$\dots$}\\\hline
\end{tabular}

\vspace{.3cm}                        

\lform{Inn1:} runs in innings, \lform{TR:} team runs, \lform{TH:}  team hits, \lform{E:}  errors,
 \lform{AB:} at-bats, \lform{RBI:} runs-batted-in,  \lform{BR:} batter runs, 
\lform{BH:} batter hits, \lform{H/V:} home or visiting,
  \lform{W:} wins, \lform{L:} losses,  \lform{IP:} innings pitched, \lform{PH:} hits given, 
\lform{PR:} runs given, \lform{ER:} earned runs,
  \lform{BB:} walks, \lform{K:} strike outs, \lform{INN:} inning with (T)op/(B)ottom,
\lform{PL-ID:} play id.

\end{minipage}
\begin{minipage}{0.5\textwidth}
\vspace*{-2.5cm}
\footnotesize
\hspace*{-1.6cm}\begin{tabular}{@{}p{10cm}@{}} 
\multicolumn{1}{@{}l@{}}{(C)} \\
\hline 
  
                  {\scriptsize
                  KANSAS CITY, Mo. -- \textcolor{olive}{Brad Keller kept up his recent
                  pitching surge with another strong
                  outing.} \textcolor{blue}{\xml{P}}
		  \textcolor{cyan}{
                  Keller gave up a home run to the first batter of the game -- Cedric Mullins -- but quickly settled in to pitch eight strong innings in the Kansas City Royals' 9--2 win over the Baltimore Orioles in a matchup of the teams with the worst records in the majors.} \textcolor{blue}{\xml{P}}
                  \textcolor{magenta}{Keller (7--5) gave up two runs and four hits with two walks and four strikeouts to improve to 3--0 with a 2.16 ERA in his last four starts}. \textcolor{blue}{\xml{P}}
                  \textcolor{teal}{Ryan O'Hearn homered among his three hits and drove in four runs, Whit Merrifield scored three runs, and Hunter Dozier and Cam Gallagher also went deep to help the Royals win for the fifth time in six games on their current homestand.} \textcolor{blue}{\xml{P}}
		  \textcolor{red}{
                  With the score tied 1--1 in the fourth, Andrew Cashner (4--13) gave up a sacrifice fly to Merrifield after loading the bases on two walks and a single. Dozier led off the fifth inning with a 423-foot home run to left field to make it 3-1.} \textcolor{blue}{\xml{P}}
                  \textcolor{orange}{The Orioles pulled within a run in the sixth when Mullins led off with a double just beyond the reach of Dozier at third, advanced to third on a fly ball and scored on Trey Mancini's sacrifice fly to the wall in right.} \textcolor{blue}{\xml{P}} \lform{$\dots$}
}\\ \hline
\end{tabular}
\end{minipage}
\begin{minipage}{\textwidth}
\vspace{-2.5cm}

\hspace*{6.7cm}
\footnotesize
\begin{tabular}{@{}l@{~~}c@{~~}c@{~~}c@{~~}cr@{~~}c@{~~}c@{~~}l@{}} 
\multicolumn{9}{@{}l@{}}{(B)} \\
\hline
\lform{BATTER}&\lform{PITCHER}&\lform{SCORER}&\lform{ACTION}&\lform{TEAM} & \lform{INN} & \lform{PL-ID} & \lform{SCORE} & \lform{$\dots$}\\ \hline
\lform{C.Mullins}&\lform{B.Keller}&\lform{-}&\lform{Home run}&\lform{Orioles} & \lform{1-T} &  \lform{1} &  \lform{1} & \lform{$\dots$}\\ 
\lform{H.Dozier}&\lform{A.Cashner}&\lform{W.Merrifield}&\lform{Grounded}&\lform{Royals} &  \lform{1-B} &\lform{3} &  \lform{1} & \lform{$\dots$}\\ 
\lform{W.Merrifield}&\lform{A.Cashner}&\lform{B.Goodwin}&\lform{Sac fly}&\lform{Royals} & \lform{4-B} &  \lform{5}& \lform{2} & \lform{$\dots$}\\ 
\lform{H.Dozier}&\lform{A.Cashner}&\lform{-}&\lform{Home run}&\lform{Royals} & \lform{5-B}&  \lform{1} & \lform{3} & \lform{$\dots$}\\ 
\lform{$\dots$} & \lform{$\dots$} & \lform{$\dots$} & \lform{$\dots$} &  \lform{$\dots$} & \lform{$\dots$} & \lform{$\dots$} & \lform{$\dots$}\ & \lform{$\dots$}\\\hline
\end{tabular}
\end{minipage}

\raisebox{.83cm}[0pt]{(D)}
\begin{minipage}{\textwidth}
\begin{center}
\footnotesize
\fontfamily{cmtt}\selectfont\footnotesize
 \begin{tabular}{p{0.32\columnwidth}} \\\hline 
    V(Orioles),    V(Royals),\\
   V(C.Mullins),
    V(J.Villar), 
    V(W.Merrifield),
    V(R.O'Hearn),
    V(A.Cashner), 
    V(B.Keller), 
    V(H.Dozier),
    $\dots$, \\ 
V(1-T), 
    V(1-B), 
    V(2-T),
    V(2-B), 
    V(3-T), 
    V(3-B),
$\dots$ \\ \hline
  \end{tabular}
\quad
  \begin{tabular}{p{0.5\columnwidth}} \\\hline 
    V(Royals) V(Orioles), \\
V(Orioles) V(C.Mullins), V(Orioles) V(J.Villar), 
V(Royals) V(W.Merrifield), 
V(Royals) V(R.O'Hearn),  
V(Orioles) V(A.Cashner),
V(Royals) V(B.Keller), $\dots$,\\
V(C.Mullins) V(Royals) V(Orioles), \\
V(J.Villar) V(Royals) V(Orioles),
$\dots$
 \\\hline
  \end{tabular}
  \end{center}
 \end{minipage}

(E)\hspace*{.3cm}\begin{minipage}{\textwidth}
\begin{center}
\footnotesize
  \begin{tabular}{@{}p{14.5cm}@{}} 
   \multicolumn{1}{c}{}\\\hline
\multicolumn{1}{c}{\vspace*{-.2cm}} \\
   {\fontfamily{cmtt}\selectfont\footnotesize

    \textcolor{olive}{V(B.Keller)}\textcolor{blue}{\xml{P}}\textcolor{cyan}{V(B.Keller) V(C.Mullins)
    V(Royals)
    V(Orioles)}\textcolor{blue}{\xml{P}}\textcolor{magenta}{V(B.Keller)}\textcolor{blue}{\xml{P}}  \textcolor{teal}{V(R.O'Hearn)
    V(W.Merrifield) V(H.Dozier) V(C.Gallagher)} \textcolor{blue}{\xml{P}}\textcolor{red}{V(4-B, 5-B)} \textcolor{blue}{\xml{P}}
    \textcolor{orange}{V(6-T)}\textcolor{blue}{\xml{P}}}\\ \hline
  \end{tabular}
  \end{center}
   \end{minipage}
   \caption{MLB statistics tables and game summary. Tables summarize
     the performance of teams and individual team members who played
     as batters and pitchers as well as the most important actions
     (and their actors) in each play (Tables~(A) and (B)). Macro plan
     for the game summary is shown at the bottom (Table~(E)).
     \textcolor{blue}{\xml{P}} indicates paragraph delimiters. There
     is a plan for every paragraph in the game summary (correspondence
     shown in same color); \xml{V(entity)} verbalizes entities, while
     \xml{V(inning-T/B)} verbalizes events related to the top/bottom
     side of an inning (see Section~\ref{par:plan-structure}). Set of
     candidate paragraph plans are shown above macro plan (Table~(D))
     and grouped into two types: plans describing a single
     entity/event or their combinations.  {Best viewed in color.}}
\label{fig:example}
\end{figure*}

Traditional methods for data-to-text generation
\cite{P83-1022,mckeown1992text,Reiter:1997:BAN:974487.974490} follow a
pipeline architecture, adopting separate stages for \emph{text
  planning} (determining which content to talk about and how it might
be organized in discourse), \emph{sentence planning} (aggregating
content into sentences, deciding specific words to describe concepts
and relations, and generating referring expressions), and
\emph{linguistic realisation} (applying the rules of syntax,
morphology and orthographic processing to generate surface forms).
Recent neural network-based approaches \cite{lebret-etal-2016-neural,
  mei-etal-2016-talk,wiseman-etal-2017-challenges} make use of the
encoder-decoder architecture \cite{sutskever2014sequence}, are trained
end-to-end, and have no special-purpose modules for how to best
generate a text, aside from generic mechanisms such as attention and
copy
\cite{DBLP:journals/corr/BahdanauCB14,gu-etal-2016-incorporating}. The
popularity of end-to-end models has been further boosted by the
release of new datasets with thousands of input-document training
pairs. The example shown in Figure~\ref{fig:example} is taken from the
MLB dataset \cite{puduppully-etal-2019-data} which contains baseball
game statistics and human written summaries (\textasciitilde
25K~instances).  \textsc{RotoWire} \cite{wiseman-etal-2017-challenges}
is another widely used benchmark, which contains NBA basketball game
statistics and their descriptions (\textasciitilde 5K~instances).

Despite being able to generate fluent text, neural data-to-text
generation models are often imprecise, prone to hallucination (i.e.,
generate text that is not supported by the input), and
poor at content selection and document structuring
\cite{wiseman-etal-2017-challenges}.
Attempts to remedy some of these issues focus on changing the way
entities are represented
\cite{puduppully-etal-2019-data,iso-etal-2019-learning}, allowing the
decoder to skip low-confidence tokens to enhance faithful generation
\cite{DBLP:journals/corr/abs-1910-08684}, and making the encoder-decoder architecture
more modular by introducing \emph{micro planning}
\cite{DBLP:journals/corr/abs-1809-00582,
  moryossef-etal-2019-step}. Micro planning operates at the record
level (see Tables~(A) Figure~\ref{fig:example}; e.g., \lform{\small
  C.Mullins BH 2}, \lform{\small J.Villar TEAM Orioles}), it
determines which facts should be mentioned within a textual unit
(e.g.,~a sentence) and how these should be structured (e.g.,~the
sequence of records). An explicit content planner essentially makes
the job of the neural network less onerous allowing to concentrate on
producing fluent natural language output, without expending too much
effort on content organization.

In this work, we focus on \emph{macro planning}, the high-level
organization of information and how it should be presented which we
argue is important for the generation of long, multi-paragraph
documents (see text~(C) in Figure~\ref{fig:example}). Problematically,
modern datasets like MLB (\citealt{puduppully-etal-2019-data}; and
also Figure~\ref{fig:example}) and \textsc{RotoWire}
\cite{wiseman-etal-2017-challenges} do not naturally lend themselves
to document planning as there is no explicit link between the summary
and the content of the game (which is encoded in tabular form). In
other words, the underlying plans are \emph{latent}, and it is not
clear how they might be best represented, i.e., as sequences of
records from a table, or simply words. Nevertheless, game summaries
through their segmentation into paragraphs (and lexical overlap with
the input) give clues as to how content might be organized. Paragraphs
are a central element of discourse
\cite{Chafe:1979,Longacre:1979,Halliday:Hasan:76}, the smallest domain
where coherence and topic are defined and anaphora resolution is
possible \cite{zadrozny-jensen-1991-semantics}. We therefore
operationalize the macro plan for a game summary as a \emph{sequence}
of \emph{paragraph} plans.

Although resorting to paragraphs describes the summary plan at a
{coarse} level, we still need to specify individual paragraph
plans. In the sports domain, paragraphs typically mention entities
(e.g,~players important in the game), key events (e.g.,~scoring a
run), and their interaction. And most of this information is
encapsulated in the statistics accompanying game summaries (see
Tables~(A) and (B) in Figure~\ref{fig:example}). We thus define
paragraph plans such that they contain verbalizations of entity and
event records (see plan~(E) in Figure~\ref{fig:example}). Given a
\emph{set} of paragraph plans and their corresponding game summary
(see Tables~(D) and summary~(C) in Figure~\ref{fig:example}), our task
is twofold. At training time, we must \emph{learn} how content was
selected in order to give rise to specific game summaries (e.g., how
input~(D) led to plan~(E) for summary (C) in
Figure~\ref{fig:example}), while at test time, given input for a new
game, we first predict a macro plan for the summary and then generate
the corresponding document.

We present a two-stage approach where macro plans are induced from
training data (by taking the table and corresponding summaries into
account) and then fed to the text generation stage. Aside from making
data-to-text generation more interpretable, the task of generating a
document from a macro plan (rather than a table) affords greater
control over the output text and plays to the advantage of
encoder-decoder architectures which excel at modeling sequences.  We
evaluate model performance on the \textsc{RotoWire}
\cite{wiseman-etal-2017-challenges} and MLB
\cite{puduppully-etal-2019-data} benchmarks. Experimental results show
that our plan-and-generate approach produces output which is more
factual, coherent, and fluent compared to existing state-of-the-art
models. Our code, trained models and dataset with
macro plans can be found at 
\url{https://github.com/ratishsp/data2text-macro-plan-py}.

\section{Related Work}
\label{sec:related-work}

Content planning has been traditionally considered a fundamental
component in natural language generation. Not only does it determine
which information-bearing units to talk about, but also arranges them
into a structure that creates coherent output. Many content planners
have been based on theories of discourse coherence
\cite{hovy1993automated}, schemas \cite{mckeown-etal-1997-language} or
have relied on generic planners \cite{dale1989generating}.  Plans are
mostly based on hand-crafted rules after analyzing the target text,
although a few approaches have recognized the need for learning-based
methods. For example, \citet{duboue-mckeown-2001-empirically} learn
ordering constraints in a content plan,
\citet{konstas-lapata-2013-inducing} represent plans as grammar rules
whose probabilities are estimated empirically, while others make use
of semantically annotated corpora to bootstrap content planners
\cite{duboue-mckeown-2002-content,kan-mckeown-2002-corpus}.

More recently, various attempts have been made to improve neural
generation models \cite{wiseman-etal-2017-challenges} based on the
encoder-decoder architecture \cite{DBLP:journals/corr/BahdanauCB14} by
adding various planning
modules. \citet{DBLP:journals/corr/abs-1809-00582} propose a model for
data-to-text generation which first learns a plan from the
records in the input table and then generates a summary conditioned on
this plan. \citet{shao-etal-2019-long} introduce a Planning-based
Hierarchical Variational Model where a plan is a sequence of groups,
each of which contains a subset of input items to be covered in a
sentence. The content of each sentence is verbalized, conditioned on
the plan and previously generated context. In their case, input items
are a relatively small list of attributes (\textasciitilde 28) and the
output document is also short (\textasciitilde 110 words).

There have also been attempts to incorporate neural modules in a
pipeline architecture for data-to-text generation.
\citet{moryossef-etal-2019-step} develop a model with a symbolic text
planning stage followed by a neural realization stage.  They
experiment with the WebNLG dataset \cite{gardent-etal-2017-creating}
which consists of RDF $\langle$ Subject, Object, Predicate~$\rangle$
triples paired with corresponding text.  Their document plan is a
sequence of sentence plans which in turn determine the division of
facts into sentences and their order.
Along similar lines, \citet{castro-ferreira-etal-2019-neural} propose an
architecture comprising of multiple steps including discourse ordering,
text structuring, %
lexicalization,
referring expression generation, %
and surface realization. %
Both approaches show the effectiveness of pipeline architectures,
however, their task does not require content selection and the output
texts are relatively short (24 tokens on average).

Although it is generally assumed that task-specific parallel data is
available for model training,
\citet{doi:10.1162/colia00363} do away with this assumption
and present a three-stage pipeline model which learns from monolingual
corpora. They first convert the input to a form of tuples, which in
turn are expressed in simple sentences, followed by the third stage of
merging simple sentences to form more complex ones by aggregation and
referring expression generation. They also evaluate on data-to-text
tasks which have relatively short outputs.  There have also been
efforts to improve the coherence of the output, especially when
dealing with longer documents.  \citet{puduppully-etal-2019-data} make
use of hierarchical attention over entity representations which are
updated dynamically, while \citet{iso-etal-2019-learning} explicitly
keep track of salient entities and memorize which ones have been
mentioned.

Our work also attempts to alleviate deficiencies in neural
data-to-text generation models. In contrast to previous approaches,
\cite{DBLP:journals/corr/abs-1809-00582,moryossef-etal-2019-step,doi:10.1162/colia00363},
we place emphasis on \emph{macro planning} and create plans
representing high-level organization of a document including both its
content \emph{and} structure. We share with previous work (e.g.,
\citealt{moryossef-etal-2019-step}) the use of a two-stage architecture.
We show that macro planning can be
successfully applied to \emph{long} document data-to-text generation
resulting in improved factuality, coherence, and fluency without any
postprocessing (e.g., to smooth referring expressions) or recourse to
additional tools (e.g.,~parsing or information extraction).

\section{Problem Formulation}

We hypothesize that generation based on plans should fare better
compared to generating from a set of records, since macro plans offer
a bird's-eye view, a high-level organization of the document content
and structure.  %
We also believe that macro planning will work well for long-form text
generation, i.e., for datasets which have multi-paragraph target
texts, a large vocabulary space, and require content selection.

We assume the input to our model is a set of paragraph plans
$\mathcal{E}=\{e_i\}_{i=1}^{|\mathcal{E}|}$ where $e_i$ is a paragraph 
plan.  We model the process of generating output
summary~$y$ given~$\mathcal{E}$ as a two step process, namely the construction
of a macro plan~$x$ based on the set of paragraph plans, followed by the
generation of a summary given a macro plan as input.  We now explain
how~$\mathcal{E}$ is obtained and each step is realized. \label{sec:macro-plan}
We discuss our model considering mainly an example from the MLB
dataset \cite{puduppully-etal-2019-data} but also touch on how the
approach can be straightforwardly adapted to \textsc{RotoWire}
\cite{wiseman-etal-2017-challenges}.

\subsection{Macro Plan Definition}
\label{par:plan-structure}

A macro plan consists of a sequence of paragraph plans separated by a
paragraph discourse marker \textcolor{blue}{\xml{P}},  i.e.,
{$x= e_i$} \textcolor{blue}{\xml{P}} $e_j \dots $\textcolor{blue}{\xml{P}} 
{$e_k$ where $e_i, e_j, e_k \in \mathcal{E}$.}  A paragraph
plan in turn is a sequence of entities and events describing the
game. By \emph{entities} we mean individual players or teams and the
information provided about them in box score statistics (see rows and
column headings in Figure~\ref{fig:example} Table~(A)), while
\emph{events} refer to information described in play-by-play (see
Table~(B)). In baseball, plays are grouped in half-innings. During each
half of an inning, a team takes its turn to bat (the visiting team
bats in the top half and the home team in the bottom half).
An example macro plan is shown at the bottom of
Figure~\ref{fig:example}.  Within a paragraph plan, entities and
events are verbalized into a \emph{text sequence} along the lines of
\citet{saleh-etal-2019-naver}. We make use of special tokens for the
\xml{TYPE} of record followed by the value of record from the
table. We retain the same position for each record type and
value. %
For example, batter \textsl{C.Mullins} from Figure~\ref{fig:example}
would be verbalized as \xml{PLAYER}\cmtt{C.Mullins} \xml{H/V}\cmtt{H}
\xml{AB}\cmtt{4} \xml{BR}\cmtt{2} \xml{BH}\cmtt{2} \xml{RBI}\cmtt{1}
\xml{TEAM}\cmtt{Orioles} $\dots$. For the sake of brevity we use
shorthand \xml{V(C.Mullins)} for the full entity.

\paragraph{Paragraph Plan for Entities}
For a paragraph containing entities, the corresponding plan will be a
verbalization of the entities in sequence.  For paragraphs with
multiple mentions of the same entity, the plan will verbalize an
entity only once and at its first position of mention.  Paragraph
``\textcolor{cyan}{\textsl{Keller gave up a home run $\dots$ the teams
    with the worst records in the majors}}'' from the summary in
Figure~\ref{fig:example} describes four entities including
\textsl{B. Keller}, \textsl{C.  Mullins}, \textsl{Royals} and
\textsl{Orioles}. The respective plan is the verbalization of the four
entities in sequence: \textcolor{cyan}{\xml{V(B.Keller)}
  \xml{V(C.Mullins)} \xml{V(Royals)} \xml{V(Orioles)}}, where \cmtt{V}
stands for verbalization and \xml{V(B.~Keller)} is a shorthand for
\xml{PLAYER}\cmtt{B.Keller} \xml{H/V}\cmtt{V} \xml{W}\cmtt{7}
\xml{L}\cmtt{5} \xml{IP}\cmtt{8} $\dots$, \xml{V(Royals)} is a
shorthand for the team \xml{TEAM}\cmtt{Royals} \xml{TR}\cmtt{9}
\xml{TH}\cmtt{14} \xml{E}\cmtt{1}, and so on.

\paragraph{Paragraph Plan for Events}
A paragraph may also describe one or more events. For example, the
paragraph ``\textcolor{red}{\textsl{With the score tied 1--1 in the
fourth}} \textcolor{red}{\textsl{$\dots$ 423-foot home run to left field 
to make it 3-1}}''
discusses what happened in the bottom halves of the \textsl{fourth}
and \textsl{fifth} innings.  We verbalize an event by first describing
the participating entities followed by the plays in the event.
Entities are described in the order in which they appear in a play,
and within the same play we list the batter followed by the pitcher,
fielder, scorer, and basemen. The paragraph plan corresponding to the
bottom halves of the \textsl{fourth} and \textsl{fifth} inning is
\mbox{\textcolor{red}{\xml{V(4-B, 5-B)}}}. Here,
\textcolor{red}{\xml{V(4-B, 5-B)}} is a shorthand for
\xml{V(W.Merrifield)} \mbox{\xml{V(A.Cashner)} \xml{V(B.Goodwin)}
  $<$\cmtt{V(H.Do}}\- \cmtt{zier)}$>$ $\dots$ \xml{V(4-B,1)}
\xml{V(4-B,2)} $\dots$ \xml{V(5-B,1)} \xml{V(5-B,2)}, and so on.  The
entities \xml{V(W.Merrifield)}, \xml{V(A.Cashner)},
\xml{V(B.Goodwin)}, and \xml{V(H.Dozier)} correspond in turn to
\textsl{W. Merrifield, A. Cashner, B. Goodwin}, and \textsl{H. Dozier}
while \xml{V(5-B,1)} refers to the first play in the bottom half of
the fifth inning (see the play-by-play table in
Figure~\ref{fig:example}) and abbreviates the following detailed plan:
\xml{INN}\cmtt{5} \xml{HALF}\cmtt{B} \mbox{\xml{BATTING}\cmtt{Royals}
  \xml{PITCHING}\cmtt{Orioles}} \xml{PL-ID}\cmtt{1}
\mbox{\xml{BATTER}\cmtt{H.Dozier} \xml{PITCHER}\cmtt{A.}\-} \mbox{\cmtt{Cashner}$>$
\xml{ACTION}\cmtt{Home-run} \xml{SCORES}} \cmtt{Royals-3-Orioles-1}, etc.

The procedure described above is not specific to MLB and can be ported
to other datasets with similar characteristics such as
\textsc{RotoWire}.  However, \textsc{RotoWire} does not provide
play-by-play information, and as a result there is no event
verbalization for this dataset.

\subsection{Macro Plan Construction} 
\label{sec:macro-plan-constr}

We provided our definition for macro plans in the previous sections,
however, it is important to note that such macro plans are not readily
available in data-to-text benchmarks like MLB
\cite{puduppully-etal-2019-data} and \textsc{RotoWire}
\cite{wiseman-etal-2017-challenges} which consist of tables of
records~$r$ paired with a gold summary~$y$ (see Tables~(A)--(C) in
Figure~\ref{fig:example}). We now describe our method for obtaining
macro plans~$x$ from $r$ and~$y$. %

Similar to \citet{moryossef-etal-2019-step}, we define macro plans to
be conformant with gold summaries such that (1)~they have the same
splits into paragraphs --- entities and events within a paragraph
in~$y$ are grouped into a paragraph plan in~$x$; and (2) the order of
events and entities in a paragraph and its corresponding plan are
identical.  We construct macro plans by matching entities and events
in the summary to records in the tables. Furthermore, paragraph
delimiters within summaries form natural units which taken together
give rise to a high-level document plan.

We match entities in summaries with entities in tables using exact
string match, allowing for some degree of variation in the expression
of team names (e.g., \textsl{A's} for \textsl{Athletics} and
\textsl{D-backs} for \textsl{Diamondbacks}).  Information pertaining
to innings appears in the summaries in the form of ordinal numbers
(e.g.,~\textsl{first}, \textsl{ninth}) modifying the noun
\textsl{inning} and can be relatively easily identified via pattern
matching (e.g., in sentences like ``\textsl{Dozier led off the
  \textbf{fifth} inning}''). However, there are instances where the
mention of innings is more ambiguous (e.g., ``\textsl{With the scored
  tied 1--1 in the \textbf{fourth}, Andrew Cashner (4-13) gave up a
  sacrifice fly}'').  We could disambiguate such mentions manually and
then train a classifier to learn to predict whether an inning is
mentioned.  Instead, we explore a novel annotation-free method which
makes use of the pretrained language model GPT2
\cite{Radford2019LanguageMA}.  Specifically, we feed the context
preceding the ordinal number to GPT2 (i.e., the current paragraph up
to the ordinal number and the paragraph preceding it) and if
\textsl{inning} appears in the top 10 next word predictions, we
consider it a positive match. On a held out dataset, this method
achieves 98\% precision and 98\% recall at disambiguating inning
mentions.

To resolve whether the summary discusses the top or bottom side of an
inning, we compare the entities in the paragraph with the entities in
each half-inning (play-by-play Table~(B) in Figure~\ref{fig:example})
and choose the side with the greater number of entity matches.  For
instance, \textsl{Andrew Cashner}, \textsl{Merrifield} and
\textsl{fourth} inning uniquely resolves to the bottom half of the
\textsl{fourth} inning.

\subsection{Paragraph Plan Construction}
\label{sec:paragr-plan-constr}

Figure~\ref{fig:example} shows the macro plan we obtain for game
summary~(C). Importantly, macro plan~(E) is the outcome of a content
selection process after considering several candidate paragraph plans
as input. So, what are the candidate paragraph plans which give rise
to macro plan~(E)?  To answer this question, we examined the empirical
distribution of paragraph plans in MLB and \textsc{RotoWire} (training
portion). Interestingly, we found that \textasciitilde 79\% of the
paragraph plans in MLB refer to a single event or a single player (and
team(s)). In \textsc{RotoWire}, \textasciitilde 92\% of paragraphs are
about a singleton player (and team(s)) or a pair of players.

Based on this analysis, we assume that paragraph plans can be either
one (verbalized) entity/event or a combination of at most two. Under
this assumption, we explicitly enumerate the set of candidate
paragraph plans in a game. For the game in Figure~\ref{fig:example},
candidate paragraph plans are shown in Tables~(D). The first table
groups plans based on individual verbalizations describing the
team(s), players, and events taking place in specific innings. The
second table groups pairwise combinations thereof. In MLB, such
combinations are between team(s) and players. In \textsc{RotoWire}, we
also create combinations between players. Such paragraph plans form
set~$\mathcal{E}$ based on which macro plan~$x$ is constructed to give rise to
game summary~$y$.

\begin{figure}[t]
  \centering
  \includegraphics[width=\linewidth]{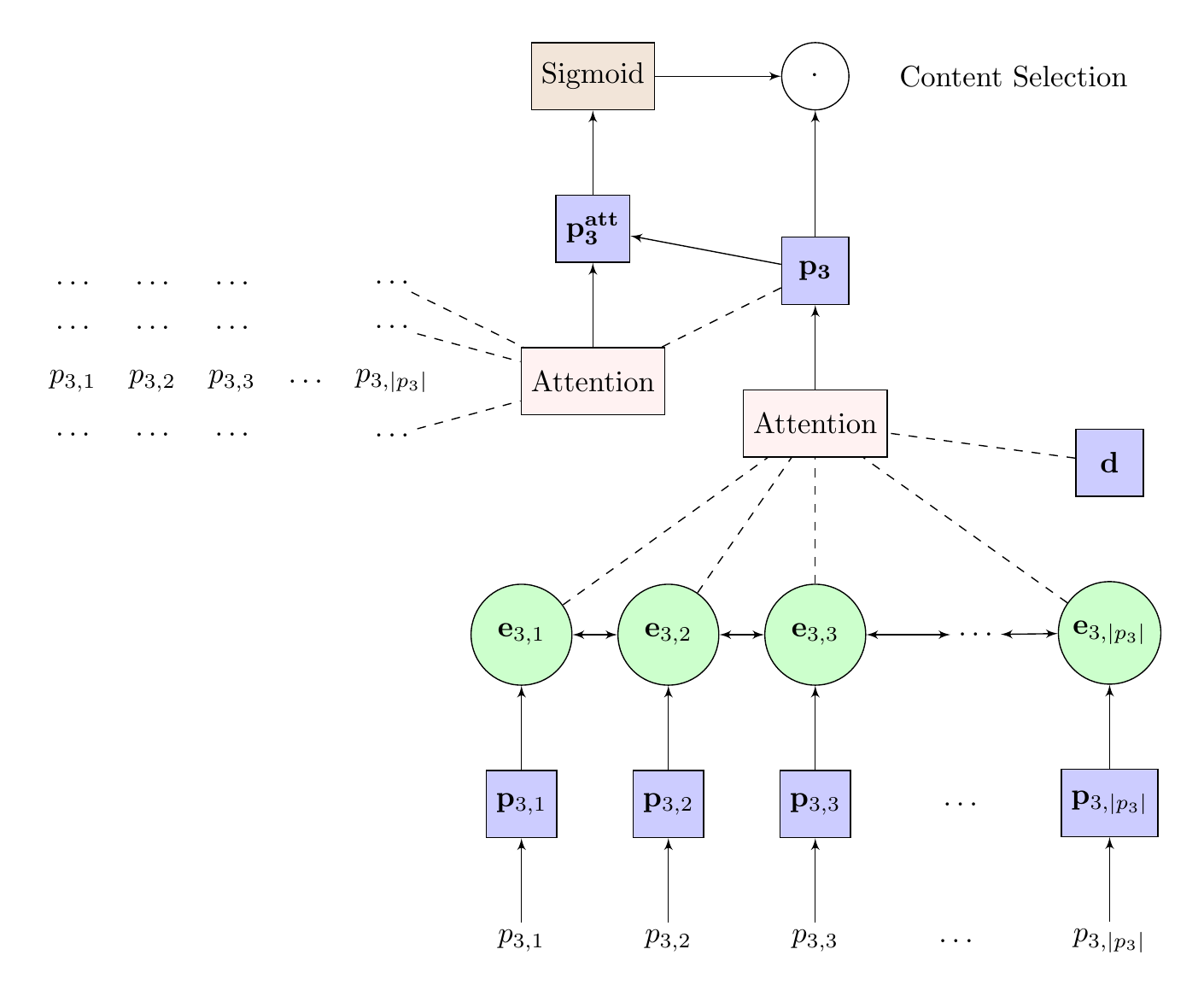}
  \caption{Paragraph plan representation and contextualization for
    macro planning.  Computation of~$\mathbf{e}_3$ is detailed in
    Equations (\ref{eq:alpha-ij}), (\ref{eq:pp-representation}),
    $\mathbf{e}_3^{att}$ in Equation~(\ref{eq:e-i-att}), and
    $\mathbf{e}_3^{c}$ in Equation~(\ref{eq:q-cs}). }
  \label{fig:content-selection}
\end{figure}

\section{Model Description}
\label{sec:model-description}

The input to our model is a set of paragraph plans each of which is a
sequence of tokens. We first compute paragraph plan representations
$\in \mathbb{R}^n$, and then apply a contextualization and content
planning mechanism similar to planning modules introduced in earlier
work \cite{DBLP:journals/corr/abs-1809-00582,chen-bansal-2018-fast}.
Predicted macro plans serve as input to our text generation model
which adopts an encoder-decoder architecture
\cite{DBLP:journals/corr/BahdanauCB14,luong-etal-2015-effective}.

\subsection{Macro Planning}
\label{sec:cs_planning}

\paragraph{Paragraph Plan Representation}
We encode tokens in a verbalized paragraph plan~$e_i$ as $\{
\mathbf{e}_{i,j} \}_{j=1}^{|e_i|}$ with a BiLSTM
(Figure~\ref{fig:content-selection} bottom part).  To reflect the fact
that some records will be more important than others, we compute an
attention weighted sum of $\{ \mathbf{e}_{i,j} \}_{j=1}^{|e_i|}$
following \citet{yang-etal-2016-hierarchical}.  Let $\mathbf{d} \in
\mathbb{R}^n$ denote a randomly initialized query vector learnt
jointly with the rest of
parameters. %
We compute attention values~$\alpha_{i,j}$ over $\mathbf{d}$ and
paragraph plan token representation $\mathbf{e}_{i,j}$: 
\begin{align}
\alpha_{i,j} &\propto \exp (\mathbf{d}^\intercal
\mathbf{e}_{i,j}) \label{eq:alpha-ij}
\end{align}
Paragraph plan vector~$\mathbf{e}_i$ is the attention weighted sum
of~$\mathbf{e}_{i,j}$ (with $\sum_{j} \alpha_{i,j} = 1$):
\begin{align}
\mathbf{e}_i &= \sum_j \alpha_{i,j} \mathbf{e}_{i,j} \label{eq:pp-representation}
\end{align}

Next, we contextualize each paragraph plan representation vis-a-vis other
paragraph plans (Figure \ref{fig:content-selection} top left part). 
First, we compute attention scores~$\beta_{i,k}$ over paragraph plan
representations to obtain %
an attentional vector~$\mathbf{e}_i^{att}$~for each:
\begin{align}
\beta_{i,k} &\propto \exp (\mathbf{e}_i^\intercal \mathbf{W}_a \mathbf{e}_k) \nonumber \\
\mathbf{c}_i &= \sum_{k \neq i} \beta_{i,k}\mathbf{e}_{k} \nonumber \\
\mathbf{e}_i^{att} &= \mathbf{W}_g [\mathbf{e}_i; \mathbf{c}_i] \label{eq:e-i-att}
\end{align}
where $\mathbf{W}_a \in \mathbb{R}^{n \times n} , \mathbf{W}_g \in \mathbb{R}^{n \times 2n}$
are parameter matrices, and $\sum_{k \neq i} \beta_{i,k} = 1$.
Then, we compute a content selection gate, and apply this gate to~$\mathbf{e}_i$
to obtain new paragraph plan representation~$\mathbf{e}_i^{c}$:
\begin{align}
\mathbf{g}_i &= \sigmoid \left( \mathbf{e}_i^{att} \right) \nonumber \\
\mathbf{e}_i^{c} &= \mathbf{g}_i \odot \mathbf{e}_i \label{eq:q-cs}
\end{align}
where $\odot$ denotes element-wise multiplication.  Thus, each element
in $\mathbf{e}_i$ is weighted by corresponding element of
$\mathbf{g}_i\in {[0,1]}^{n}$ to obtain a contextualized paragraph
plan representation $\mathbf{e}_i^{c}$.

\begin{figure}[t]
  \centering
  \includegraphics[width=\linewidth]{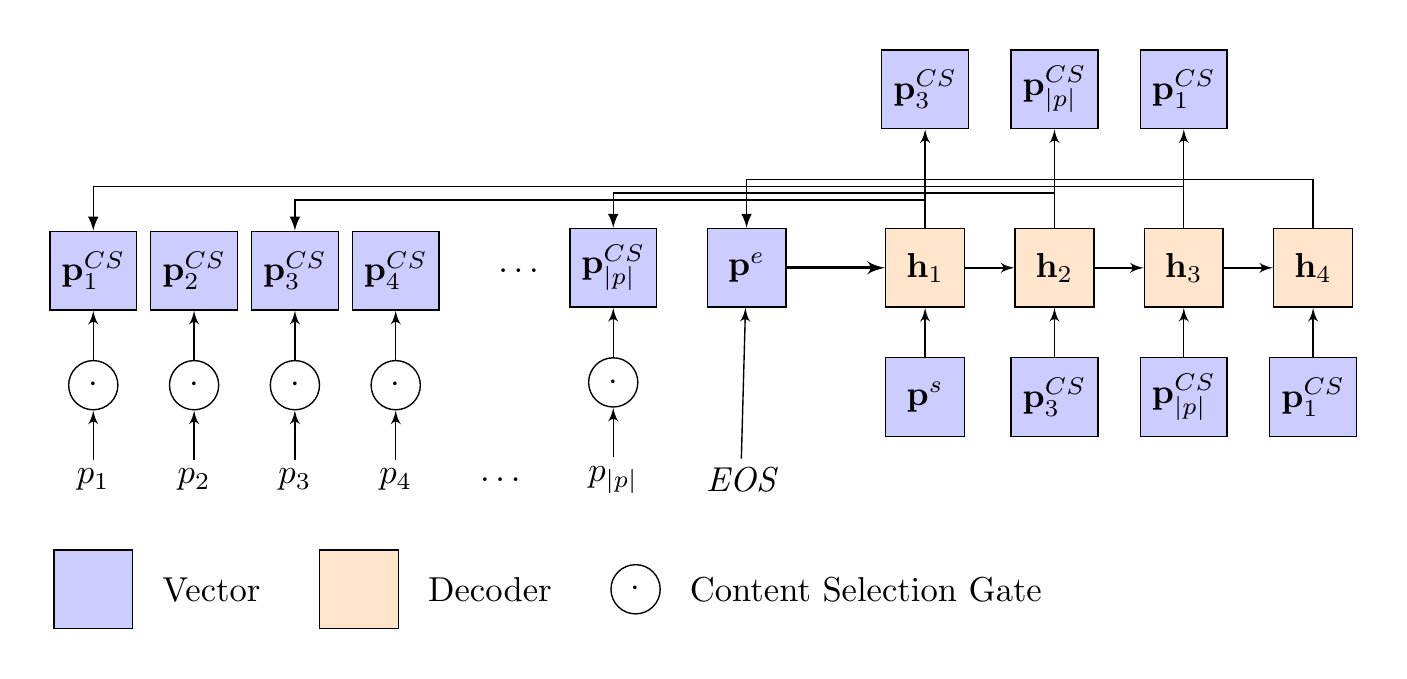} %
  \caption{Macro planning model; paragraph plan representation and
    contextualization mechanism are detailed in
    Figure~\ref{fig:content-selection}. The output points to $e_3$,
    $e_{|\mathcal{E}|}$, and $e_1$ (see Equations (\ref{eq:z-given-e}) and
    (\ref{eq:z-given-e-pointer})). \emph{EOM} is end of macro plan token.}
  \label{fig:macro-planning}
\end{figure}

\paragraph{Content Planning}
Our model learns to predict macro plans, after having been trained on
pairs of sets of paragraph plans and corresponding macro plans
(Sections~\ref{sec:macro-plan-constr} and~\ref{sec:paragr-plan-constr}
explain how we obtain these for data-to-text datasets like
\textsc{RotoWire} and MLB). More formally, we model macro plan $z =
z_1 \dots z_{|z|}$ as a sequence of pointers, with each~$z_k$ pointing
to an input paragraph plan, i.e.,~\mbox{$z_k \in \{ e_i
  \}_{i=1}^{|\mathcal{E}|}$}. We decompose $p(z|\mathcal{E})$, the probability of macro
plan~$z$ given paragraph plans~$\mathcal{E}$, as:
\begin{equation}
p(z|\mathcal{E}) = \prod_{k=1}^{|z|} {p(z_k | z_{<k},\mathcal{E})} \label{eq:z-given-e}
\end{equation}
where $z_{<k} = z_1 \dots z_{k-1}$.

We use Pointer Networks \cite{NIPS2015_5866} to model $p(z_k |
z_{<k},\mathcal{E})$ as:
\begin{align}
p(z_k = e_i | z_{<k} , \mathcal{E}) \propto \exp (\mathbf{h}_k^\intercal \mathbf{W}_b \mathbf{e}_i^{c}) \label{eq:z-given-e-pointer}
\end{align}
where $p(z_k | z_{<k},\mathcal{E})$ is normalized to 1 and $\mathbf{W}_b \in
\mathbb{R}^{n \times n}$.  Rather than computing a weighted
representation, Pointer Networks make use of attention to point to
specific elements in the input (see Figure~\ref{fig:macro-planning}).
We use a decoder LSTM to compute hidden representation $\mathbf{h}_k$
at time step~$k$. We initialize~$\mathbf{h}_0$ with the mean paragraph
plan representation, $\avg(\{ \mathbf{e}_i^{c} \}_{i=1}^{|\mathcal{E}|} )$.
Once the output points to~$e_i$, its representation~$\mathbf{e}_i^{c}$
is used as input to the next step of the LSTM decoder.  The process
stops when the model points to {\it EOM}, a token indicating end of
the macro plan.

\subsection{Text Generation}
\label{sec:text-generation}

Recall that~$z$ is a sequence of pointers with each entry~$z_k$
pointing to a paragraph plan i.e.,~$z_k\in \{ e_i\}_{i=1}^{|\mathcal{E}|}$.  We
can deterministically obtain macro plan~$x$ from~$z$ by retrieving the
paragraph plans being pointed to, adding \textcolor{blue}{\xml{P}}
 separators in between. %
The conditional output probability $p(y|x)$ is modeled as:
\begin{equation}
p(y|x) = \prod_{t=1}^{|y|} p(y_t|y_{<t},x) \nonumber
\end{equation}
where $y_{<t} = y_1 \dots y_{t-1}$.

To compute~$p(y|x)$, we use an encoder-decoder architecture enhanced
with an attention mechanism
\cite{DBLP:journals/corr/BahdanauCB14,luong-etal-2015-effective}. We
encode macro plan~$x$ with a bidirectional LSTM
\cite{hochreiter1997long}.
At time step~$t$, we lookup the embedding of the previously predicted
word~$y_{t-1}$ and feed it as input to the decoder which is another
LSTM
unit. %
The decoder attends over hidden states of the macro plan to
predict~$y_{t}$. We further incorporate a copy mechanism
\cite{gulcehre-etal-2016-pointing} in the decoder to enable copying
values directly from the macro plan.

We expect the text generation model to learn to generate summary
tokens while focusing on the corresponding macro plan and that the
output summary will indeed follow the plan in terms of the entities
and events being described and their
order.  %
At the same time, we believe that text generation is relatively easier
as the encoder-decoder model is relieved from the tasks of document
structuring and information selection.

\subsection{Training and Inference}
We train two independent models for macro planning and text 
generation.
Our training objective for macro planning aims to maximize the log 
likelihood of the
macro plan given the paragraph plans:
\begin{equation}
\max_\theta \sum_{(\mathcal{E}, z) \in \mathcal{D} }{ \log{p \left( z | \mathcal{E}; \theta \right)}} \nonumber
\end{equation}
where $\mathcal{D}$ is the training set consisting of pairs of (sets
of) paragraph plans and macro plans, and $\theta$ are model
parameters.%

Our training objective for text generation aims to maximize the log 
likelihood of the output text given the macro plan:
\begin{equation}
\max_\phi \sum_{(x, y) \in \mathcal{F} }{ \log{p \left( y | x; \phi \right)} } \nonumber
\end{equation}
where $\mathcal{F}$ is the training set consisting of pairs of macro plans 
and game summaries, and $\phi$ are model parameters.

During inference, we employ beam search to find the most likely 
macro plan $\hat{z}$ among candidate macro plans $z'$ given paragraph plans 
as input.
\begin{align}
\hat{z} &= \argmax_{z'} p(z' | \mathcal{E}; \theta) \nonumber
\end{align}
We deterministically obtain $\hat{x}$ from $\hat{z}$, and output summary
$\hat{y}$~among candidate outputs $y'$ given macro plan~$\hat{x}$ as
input:
\begin{align}
\hat{y} &= \argmax_{y'} p(y' | \hat{x}; \phi) \nonumber
\end{align}

\section{Experimental Setup}
\label{sec:experimental-setup}

\begin{table}[t]
  \footnotesize
\centering
  \begin{tabular}{lcc} \thickhline 
  & \textsc{RotoWire} & MLB \\ 
\thickhline 
Vocab Size & 11.3K & 38.9K \\ 
\# Tokens & 1.5M & 14.3M \\ 
\# Instances & 4.9K & 26.3K \\ 
\# Record Types & 39 & 53 \\ 
Avg Records & 628 & 565 \\ 
Avg Paragraph Plans & 10.7 & 15.1 \\ 
Avg  Length & 337.1 & 542.05 \\ \thickhline  
\end{tabular}
\caption{\label{dataset-stats}
Dataset statistics for  \textsc{RotoWire} and MLB.  Vocabulary size,
number of tokens, number of instances (i.e., table-summary pairs),
number of record types, average number of records, average number of
paragraph plans, and 
average summary length.}
    \end{table}

\paragraph{Data}

We performed experiments on the \textsc{RotoWire}
\cite{wiseman-etal-2017-challenges} and MLB
\cite{puduppully-etal-2019-data} benchmarks.  The details of these two
datasets are given in Table~\ref{dataset-stats}. We can see that MLB
is around 5 times bigger, has a richer vocabulary and longer game
summaries.  We use the official splits of 3,398/727/728 for
\textsc{RotoWire} and 22,821/1,739/1,744 for MLB.
We make use of
a tokenization script\footnote{\url{https://github.com/neulab/DGT}} to
detokenize and retokenize the summaries in both \textsc{RotoWire} and MLB.

We reconstructed the MLB dataset, as the version released by
\citet{puduppully-etal-2019-data} had removed all paragraph delimiters
from game summaries. Specifically, we followed their methodology and
downloaded the same summaries from the ESPN
website\footnote{\url{http://www.espn.com/mlb/recap?gameId={gameid}}}
and added the \textcolor{blue}{\xml{P}} delimiter to paragraphs in the
summaries.\footnote{Although our model is trained on game summaries
  with paragraph delimiters, and also predicts these at generation
  time, for evaluation we strip \textcolor{blue}{\xml{P}} from model
  output.}
\textsc{RotoWire} does not have paragraph delimiters in game summaries
either. We reverse engineered these as follows: (1)~we split summaries
into sentences using the NLTK \cite{BirdKleinLoper09} sentence
tokenizer; (2) initialized each paragraph with a separate sentence;
(3) merged two paragraphs into one if the entities in the former were
a superset of entities in the latter; (4) repeated Step~3 until no
 merges were possible.

\paragraph{Training Configuration}
We tuned the model hyperparameters on the development set. For
training the macro planning and the text generation stages, we used
the Adagrad \cite{DBLP:journals/jmlr/DuchiHS11}
optimizer. Furthermore, the text generation stage made use of
truncated BPTT \cite{DBLP:journals/neco/WilliamsP90} with truncation
length~100.  We learn subword vocabulary
\cite{sennrich-etal-2016-neural} for paragraph plans in the macro
planning stage. We used 2.5K merge operations for \textsc{RotoWire}
and 8K merge operations for MLB.  In text generation, we learn a joint
subword vocabulary for the macro plan and game summaries.  We used 6K
merge operations for \textsc{RotoWire} and 16K merge operations for
MLB. All models were implemented on OpenNMT-py \cite{klein-etal-2017-opennmt}.  We
add to set~$\mathcal{E}$ the paragraph plans corresponding to the
output summary paragraphs, to ensure full coverage during training of
the macro planner.  During inference for predicting macro plans, we 
employ length normalization
\cite{DBLP:journals/corr/BahdanauCB14} to avoid penalizing longer
outputs; specifically, we divide the scores of beam search by the
length of the output. In addition, we adopt bigram blocking
\cite{paulus2018a}. For MLB, we further block beams containing more
than two repetitions of a unigram. This helps improve the diversity of
the predicted macro plans.

\paragraph{System Comparisons}
We compared our model against the following systems: (1)~the
\textbf{Templ}ate-based generators from
\citet{wiseman-etal-2017-challenges} for \textsc{RotoWire} and
\citet{puduppully-etal-2019-data} for MLB. Both systems apply the same
principle, they emit a sentence about the teams playing in the game,
followed by player-specific sentences, and a closing sentence. MLB
additionally contains a description of play-by-play; (2)
\textbf{ED$+$CC}, the best performing system in
\citet{wiseman-etal-2017-challenges}, is a vanilla encoder-decoder
model equipped with an attention and copy
mechanism; %
(3) \textbf{NCP$+$CC}, the micro planning model of
\citet{DBLP:journals/corr/abs-1809-00582}, generates content plans
from the table by making use of Pointer networks \cite{NIPS2015_5866}
to point to records; content plans are encoded with a BiLSTM and the
game summary is decoded using another LSTM with attention and copy;
(4)~\textbf{ENT}, the entity-based model of
\citet{puduppully-etal-2019-data}, creates dynamically updated
entity-specific representations; the text is generated conditioned on
the data input and entity memory representations using hierarchical
attention at each time step.

\section{Results}
\label{sec:results}

\paragraph{Automatic Evaluation}

For automatic evaluation, following earlier work
(\citealt{wiseman-etal-2017-challenges,DBLP:journals/corr/abs-1809-00582,puduppully-etal-2019-data},
inter alia) we report BLEU \cite{papineni-etal-2002-bleu} with
the gold summary as reference but also make use of the Information
Extraction (IE) metrics from \citet{wiseman-etal-2017-challenges}
which are defined over the output of an IE system; the latter extracts
entity (players, teams) and value (numbers) pairs in a summary, and
then %
predicts the \textsl{type} of relation. For instance, given the pair
\textsl{Kansas City Royals, 9}, it would predict their relation as
\textsl{\lform{\normalsize TR}} (i.e.,~Team Runs).  Training data for
the IE system is obtained by checking for matches between
\textsl{entity, value} pairs in the gold summary and
\textsl{entity, value, record type} triplets in the table.

Let $\hat{y}$ be the gold summary and $y$ the model
output. \textsl{Relation Generation} (RG) measures the precision and
count of relations extracted from $y$ that also appear in
records~$r$. \textsl{Content Selection} (CS) measures the precision
and recall of relations extracted from $y$ that are also extracted
from~$\hat{y}$.  \textsl{Content Ordering} (CO) measures the
normalized Damerau-Levenshtein distance between the sequences of
relations extracted from $y$ and~$\hat{y}$.

We reused the IE model from \citet{DBLP:journals/corr/abs-1809-00582}
for \textsc{RotoWire} but retrained it for MLB to improve its
precision and recall. Furthermore, the implementation of
\citet{wiseman-etal-2017-challenges} computes RG, CS, and CO excluding
duplicate relations. This artificially inflates the performance of
models whose outputs contain repetition. We include duplicates in the
computation of the IE metrics (and recreate them for all comparison
systems).

\begin{table}[t]
\footnotesize
\centering
\begin{tabular}{@{}l@{~}|@{~}c@{~~~}c@{~}|c@{~~~}c@{~~~}c@{~}|@{~}c@{~}|@{~}c@{}} 
 \thickhline
 \multirow{2}{*}{\textsc{RotoWire}} &\multicolumn{2}{c|}{RG} &\multicolumn{3}{c@{~}|@{~}}{CS} & CO & \multirow{2}{*}{BLEU}\\ 
 &\# & P\% & P\% & R\% & F\% & DLD\% & \\ \thickhline

Templ &\textbf{54.3} &\textbf{99.9} &27.1 &{57.7} & 36.9 &13.1 &8.46  \\
WS-2017 & 34.1 & 75.1 & 20.3 & 36.3  &26.1 & 12.4 & 14.19 \\
ED$+$CC & 35.9 & 82.6 &19.8 & 33.8 & 24.9 & 12.0 & 14.99 \\ %
NCP$+$CC &{40.8} & {87.6} & 28.0 & {51.1} & 36.2
 &15.8 & {16.50} \\ 
ENT&32.7 &91.7 & \textbf{34.7} & 48.5 & 40.5 & {16.6} & 16.12 \\
RBF-2020 & 44.9 & 89.5 & 23.9 & 47.0 & 31.7 & 14.3 & \textbf{17.16} \\ \hline 
Macro & 42.1 & 97.6 & 34.1 & \textbf{57.8} & \textbf{42.9} & \textbf{17.7} & 15.46 \\
~$-$Plan(4) & 36.2 & 81.3 & 22.1 & 38.6 & 28.1 & 12.1 & 14.00 \\

\thickhline
\multicolumn{7}{c}{} \\ \thickhline
 \multirow{2}{*}{MLB} &\multicolumn{2}{c|}{RG} &\multicolumn{3}{c@{~}|@{~}}{CS} & CO & \multirow{2}{*}{BLEU}\\

 &\# & P\% & P\% & R\% & F\% & DLD\% & \\ \thickhline

Templ & \textbf{62.3} & \textbf{99.9} & 21.6 & \textbf{55.2} &  31.0 &  11.0 & 4.12 \\
ED$+$CC & 32.5 & 91.3 & 27.8 & 40.6 & 33.0 &   17.1 & 9.68 \\
NCP$+$CC& 19.6 & 81.3 & \textbf{44.5} & 44.1 & 44.3 &  \textbf{21.9} & 9.68 \\
ENT&23.8 & 81.1 & 40.9 & 49.5 &44.8  &20.7 & 11.50 \\ \hline
Macro & 30.8 & 94.4 & 40.8 & 54.9 & \textbf{46.8} & 21.8 & \textbf{12.62} \\ %
~$-$Plan(SP,4) & 25.1& 92.7 & 40.0 & 44.6 & 42.2 & \textbf{21.9} & 11.09 \\\thickhline
\end{tabular}
\caption{\label{tbl:results-with-ie-test} Evaluation on
  \textsc{RotoWire} and \textsc{MLB} test sets;  relation
  generation (RG) count (\#) and precision (P\%), content selection
  (CS) precision (P\%), recall (R\%) and F-measure (F\%), content
  ordering (CO) in normalized Damerau-Levenshtein distance (DLD\%),
  and BLEU.}
\end{table}

Table~\ref{tbl:results-with-ie-test} (top) presents our results on the
\textsc{RotoWire} test set.
In addition to Templ, NCP$+$CC, ENT, and ED$+$CC we include the best
performing model of \citet{wiseman-etal-2017-challenges} (WS-2017;
note that ED$+$CC is an improved re-implementation of their model), and
the model of \citet{rebuffel2020hierarchical} (RBF-2020) which
represents the state of the art on \textsc{RotoWire}. This model has
a Transformer encoder \cite{NIPS2017_7181} with a hierarchical
attention mechanism over entities and records within entities.
The models of \citet{saleh-etal-2019-naver},
\citet{iso-etal-2019-learning}, and \citet{gong-etal-2019-table} make
use of additional information not present in the input (e.g.,
previous/next games, summary writer) and are not directly comparable
to the systems in Table~\ref{tbl:results-with-ie-test}.  Results for
the MLB test set are in the bottom portion of
Table~\ref{tbl:results-with-ie-test}.

Templ has the highest RG precision and count on both datasets.  This
is not surprising, by design Templ is always faithful to the
input. However, notice that it achieves the lowest BLEU amongst
comparison systems indicating that it mostly regurgitates facts with
low fluency. Macro achieves the highest RG precision amongst all
neural models for \textsc{RotoWire} and MLB. We obtain an absolute
improvement of 5.9\% over ENT for \textsc{RotoWire} and 13.3\% for
MLB. In addition, Macro achieves the highest CS F-measure for both
datasets. On \textsc{RotoWire}, Macro achieves the highest CO score,
and %
the highest BLEU on MLB.  On \textsc{RotoWire}, in terms of BLEU,
Macro is worse than comparison models (e.g., NCP+CC or
ENT). Inspection of the output showed that the opening paragraph,
which mostly describes how the two teams fared, is generally shorter
in Macro, leading to shorter summaries and thus lower BLEU. There is
high variance in the length of the opening paragraph in the training
data
and Macro verbalizes the corresponding plan conservatively. Ideas such
as length normalisation \cite{DBLP:journals/corr/WuSCLNMKCGMKSJL16} or length control
\cite{kikuchi-etal-2016-controlling,takeno-etal-2017-controlling,fan-etal-2018-controllable}
could help alleviate this; however, we do not pursue them further for
fair comparison with the other models.

\paragraph{The Contribution of Macro Planning}
To study the effect of macro planning in more detail, we further
compared Macro against text generation models (see
Section~\ref{sec:text-generation}) which are trained on verbalizations
of the tabular data (and gold summaries) but do not make use of
document plans or a document planning mechanism.  On
\textsc{RotoWire}, the model was trained on verbalizations of players
and teams, with the input arranged such that the verbalization of the
home team was followed by the visiting team, the home team players and
the visiting team players. Mention of players was limited to the four
best ones, following \citet{saleh-etal-2019-naver} (see $-$Plan(4) in
Table~\ref{tbl:results-with-ie-test}).  For MLB, we additionally
include verbalizations of innings focusing on scoring plays which are
likely to be discussed in game summaries (see $-$Plan(SP,4) in
Table~\ref{tbl:results-with-ie-test}). Note that by preprocessing the
input in such a way some simple form of content selection takes place
simply by removing extraneous information which the model does not
need to consider.

Across both datasets, $-$Plan variants appear competitive. On
\textsc{RotoWire} $-$Plan(4) is better than ED$+$CC in terms of
content selection but worse compared to ENT. On MLB, $-$Plan(SP,4) is
again superior to ED$+$CC in terms of content selection but not ENT
whose performance lags behind when considering RG precision. Taken
together, these results confirm that verbalizing entities and events
into a text sequence is effective. At the same time, we see that
$-$Plan variants are worse than Macro across most metrics which underlines
the importance of an explicit planning component.

\begin{table}[t]
\footnotesize
\centering
\begin{tabular}{ lcccc }\thickhline
 Macro  & CS-P & CS-R & CS-F & CO \\ \thickhline 
\textsc{RotoWire} & 81.3 & 73.2 & 77.0 & 45.8 \\  %
MLB & 80.6 & 63.3 & 70.9& 31.4 \\
\thickhline 
\end{tabular}
\caption{Evaluation of macro planning stage; 
content selection precision (CS-P), recall
(CS-R), F-measure (CS-F) and content ordering (CO) between the inferred
plans and gold plans in terms of entities and events for
  \textsc{RotoWire} (RW) and MLB test sets.}
\label{tbl:inf-plans-cs-co}
\end{table}

Table~\ref{tbl:inf-plans-cs-co} presents intrinsic evaluation of the
macro planning stage. Here, we compare the inferred macro plan with
the gold macro plans, %
CS and CO metrics with regard to entities and events instead of
relations.  We see that our macro planning model (Macro) achieves high
scores for CS and CO for both \textsc{RotoWire} and MLB.
We further used the CS and CO metrics to check how well the generated
summary follows the (predicted) plan. We followed the steps in
Section~\ref{sec:macro-plan-constr} and reverse engineered macro plans
from the model summaries and compared these extracted plans with the
original macro plans with regard to entities and events. We found
that Macro creates summaries which follow the plan closely:
for \textsc{RotoWire}, the CS F-score and CO are greater than 98\%; 
for MLB, the CS F-score is greater than 94\% and CO
is greater than 89\%. %
We show an output summary for Macro in Table~\ref{tab:output} 
together with the predicted document
plan.

\begin{table}[t]
\scriptsize
\centering
\begin{tabular}{@{}p{7.7cm}@{}} \thickhline

\xml{V(Rays)} \textcolor{blue}{\xml{P}}
\xml{V(8-B)} \textcolor{blue}{\xml{P}}
\xml{V(Rays)} \textcolor{blue}{\xml{P}}
\xml{V(Rays)} \xml{V(Red Sox)} \textcolor{blue}{\xml{P}}
\xml{V(8-B)} \textcolor{blue}{\xml{P}}
\xml{V(8-T)} \textcolor{blue}{\xml{P}}
\xml{V(9-T)} \textcolor{blue}{\xml{P}}
\xml{V(Clay Buchholz)} \textcolor{blue}{\xml{P}}
\xml{V(5-T)} \textcolor{blue}{\xml{P}}
\xml{V(Edwin Jackson)} \textcolor{blue}{\xml{P}}
\xml{V(5-T)} \textcolor{blue}{\xml{P}}
\xml{V(8-T)}
\\ \hline 

ST. PETERSBURG, Fla. (AP) -- The \textbf{Tampa Bay Rays} are making the 
most of it. \textcolor{blue}{\xml{P}} Akinori Iwamura hit a two-run homer
in the \textbf{eighth} inning and the Rays beat the Boston Red Sox 2-1 on 
Sunday to complete a three-game sweep. \textcolor{blue}{\xml{P}} The 
\textbf{Rays}, who have the best record in the majors, have won six of their 
last seven games. \textcolor{blue}{\xml{P}} The \textbf{Rays} have won four 
of their last five series, including three in a row against the \textbf{Red 
  Sox}, who have won six of their last seven overall. \textcolor{blue}{\xml{P}} 
Dioner Navarro singled with one out in the \textbf{eighth} off Clay Buchholz 
(1-2) and moved to third on Jason Bartlett's flyout to center. Iwamura 
then drove a 1-1 pitch into the left-field stands for his second homer of the 
season. \textcolor{blue}{\xml{P}} Scott Dohmann (2-0) got the win in relief 
, striking out Manny Ramirez with runners on first and third to end the 
\textbf{eighth}. \textcolor{blue}{\xml{P}} Troy Percival worked the 
\textbf{ninth} for his fifth save in five opportunities. \textcolor{blue}
{\xml{P}} \textbf{Clay Buchholz} (1-2) gave up two runs and three hits 
in eight innings. He struck out nine and walked two. \textcolor{blue}
{\xml{P}} The Red Sox loaded the bases with one out in the \textbf{fifth} 
on a single by Coco Crisp, a wild pitch and a walk to Jed Lowrie. Jacoby 
Ellsbury drove in Crisp with a two-out single to center. \textcolor{blue}
{\xml{P}} \textbf{Jackson} struck out four and walked three. \textcolor{blue}
{\xml{P}} The Red Sox loaded the bases with one out in the \textbf{fifth} 
on a single by Coco Crisp, a walk to Jed Lowrie and a one-out walk to 
Jed Lowrie. Jackson struck out Julio Lugo, but Jacoby Ellsbury singled 
to center to put the Red Sox up 1-0. \textcolor{blue}{\xml{P}} The Red 
Sox threatened in the \textbf{eighth} when J. D. Drew drew a two-out 
walk against Trever Miller, but Ramirez struck out to end the inning. \\

\thickhline
\end{tabular}
\caption{\label{tab:output} Predicted macro plan (top) with corresponding model
  output (bottom).  Entities and events in summary corresponding to those in
  the macro plan are bold faced.}
\end{table}

\begin{table}[t]
\footnotesize
\centering
\begin{tabular}{@{}l@{~~}l@{~~}c@{~~}c@{~~}c@{~~}c@{}}
\thickhline
\textsc{RotoWire} & \#Supp & \#Contra & Gram & Coher & Concis \\
\thickhline
Gold    & 3.63 & 0.07\hspace*{0.15cm} & 38.33  & 46.25*&  30.83\\
Templ   & 7.57* & 0.08\hspace*{0.15cm} & \hspace*{-0.1cm}$-$61.67* &  \hspace*{-0.1cm}$-$52.92* & \hspace*{-0.1cm}$-$36.67*\\
ED$+$CC  & 3.92 & 0.91* & \hspace*{0.1cm}5.0 & \hspace*{-0.1cm}$-$8.33  & \hspace*{-0.1cm}$-$4.58\\
RBF-2020   & 5.08* & 0.67* & \hspace*{-0.16cm}13.33 & \hspace*{0.15cm}4.58 & \hspace*{0.15cm}3.75\\
Macro  & 4.00 & 0.27\hspace*{0.15cm} & \hspace*{-0.16cm}5.0 & \hspace*{-0.04cm}10.42&\hspace*{0.15cm}6.67\\ \thickhline
\multicolumn{6}{c}{} \\
\thickhline
MLB & \#Supp & \#Contra & Gram & Coher & Concis \\ \thickhline
Gold    & 3.59 & \hspace*{-0.1cm}0.14 & \hspace*{0.3cm}21.67\hspace*{.3cm} & \hspace*{-.25cm}30.0 & 26.67 \\
Templ   & 4.21 & \hspace*{-0.1cm}0.04 & \hspace*{-0.1cm}$-$51.25* & \hspace*{-.3cm}$-$43.75*& \hspace*{-.05cm}7.5 \\
ED$+$CC & 3.42 & \hspace*{0.05cm}0.72* & \hspace*{0.28cm}$-$22.5* \hspace*{.3cm} & \hspace*{-.4cm}$-$12.08* & \hspace*{-.05cm}$-$39.17*  \\
ENT  & 3.71 & \hspace*{0.05cm}0.73* & \hspace*{0.2cm}5.83* & \hspace*{-.2cm}$-$0.83*& \hspace*{-.05cm}$-$22.08* \\
Macro  & 3.76 & \hspace*{-.1cm}0.25 &  \hspace*{.1cm}46.25 & \hspace*{-.3cm}26.67 &27.08\\
\thickhline
\end{tabular}
\caption{Average number of supported (\#Supp) and contradicting
  (\#Contra) facts in game summaries and \textit{best-worst scaling}
  evaluation (higher is better).  Systems significantly different from
  Macro are marked with an asterisk * (using a one-way ANOVA with
  posthoc Tukey HSD tests; \mbox{$p\leq0.05$}).  }.
  \label{tbl:mlb-human-eval}
\end{table}

\paragraph{Human-Based Evaluation}
We also asked participants to assess model output in terms of relation
generation, grammaticality, coherence, and conciseness
\cite{wiseman-etal-2017-challenges,
  DBLP:journals/corr/abs-1809-00582,puduppully-etal-2019-data}, For
\textsc{RotoWire}, we compared Macro against RBF-2020\footnote{We are
  grateful to Cl\'{e}ment Rebuffel for providing us with the output of
  their system.}, ED$+$CC, Gold, and Templ. For MLB, we compared Macro
against ENT, ED$+$CC, Gold, and Templ. %
We conducted our study on the Amazon Mechanical Turk (AMT)
crowdsourcing platform,  %
following best practices for human evaluation in NLG
\cite{van-der-lee-etal-2019-best}. Specifically,
to ensure consistent ratings, we required crowdworkers to have an
approval rating greater than~98\% and a minimum of~1,000 previously
completed tasks.  Raters were restricted to English speaking countries
(i.e.,~US, UK, Canada, Ireland, Australia, or NZ). %
Participants were allowed to provide feedback on the task or field
questions (our interface accepts free text).

In our first study, we presented crowdworkers with sentences randomly
selected from summaries along with their corresponding box score (and
play-by-play in case of MLB) and asked them to count supported and
contradicting facts (ignoring hallucinations, i.e.,~unsupported
facts).  We did not require crowdworkers to be familiar with NBA or
MLB.  Instead, we provided a cheat sheet explaining the semantics of
box score tables.  In addition, we provided examples of  sentences with
supported/contradicting facts. 
We evaluated 40 summaries from the test set (20 per dataset), 4
sentences from each summary and elicited 3 responses per summary.
This resulted in 40 summaries $\times$ 5 systems $\times$ 3 raters for
a total of 600 tasks.  Altogether 131 crowdworkers participated in
this study (agreement using Krippendorff's $\alpha$ was 0.44 for
supported and 0.42 for contradicting facts).

As shown in Table~\ref{tbl:mlb-human-eval}, Macro yields the smallest
number of contradicting facts among neural models on both datasets. On
\textsc{RotoWire} the number of \emph{contradicting} facts for Macro
is comparable to Gold and Templ (the difference is not statistically
significant) and significantly smaller compared to RBF-2020 and
ED$+$CC.  The count of \emph{supported} facts for Macro is comparable
to Gold, and ED$+$CC, and significantly lower than Templ and
RBF-2020. On MLB, Macro has significantly fewer \emph{contradicting}
facts than ENT and ED$+$CC and is comparable to Templ, and Gold (the
difference is not statistically significant). The count of
\emph{supported} facts for Macro is comparable to Gold, ENT,
ED$+$CC and Templ.  For both datasets, Templ
has the lowest number of contradicting facts. This is expected as
Templ essentially parrots facts (aka records) from the table.

We also conducted a second study to evaluate the quality of the
generated summaries. We presented crowdworkers with a pair of
summaries and asked them to choose the better one in terms of
\textsl{Grammaticality} (is the summary written in well-formed
English?), \textsl{Coherence} (is the summary well structured and well
organized and does it have a natural ordering of the facts?) and
\textsl{Conciseness} (does the summary avoid unnecessary repetition
including whole sentences, facts or phrases?).  We provided example
summaries showcasing good and bad output.  For this task, we required
that the crowdworkers be able to comfortably comprehend NBA/MLB game
summaries. We elicited preferences with Best-Worst Scaling
\cite{louviere1991best, louviere2015best}, a method shown to be more
reliable than rating scales. The score of a system is computed as the
number of times it is rated best minus the number of times it is rated
worst \cite{orme2009maxdiff}. The scores range from $-$100 (absolutely
worst) to $+$100 (absolutely best).  We divided the five competing
systems into ten pairs of summaries and elicited ratings for 40
summaries (20 per dataset). Each summary pair was rated by 3
raters. This resulted in 40 summaries $\times$ 10 system pairs
$\times$ 3 evaluation criteria $\times$ 3 raters for a total of 3,600
tasks.  206 crowdworkers participated in this task (agreement using
Krippendorff's $\alpha$ was 0.47).

As shown in Table~\ref{tbl:mlb-human-eval}, on \textsc{RotoWire},
Macro is comparable to Gold, RBF-2020, and ED$+$CC in terms of
\emph{Grammaticality} but significantly better than Templ.  In terms
of \emph{Coherence}, Macro is comparable to
RBF-2020 and ED$+$CC but significantly better than Templ and 
significantly worse than Gold.
With regard to \emph{Conciseness}, Macro is comparable to Gold,
RBF-2020, and ED$+$CC, and significantly better than
Templ. %
On MLB, Macro is comparable to Gold in terms of
\emph{Grammaticality} and significantly better than ED$+$CC, ENT and Templ.
Macro is comparable to Gold in terms of \emph{Coherence} and
significantly better than ED$+$CC, ENT and Templ. In terms of
\emph{Conciseness}, raters found Macro comparable to Gold and Templ
and significantly better than ED$+$CC, and ENT.  %
Taken together, our results show that macro planning leads to improvement 
in data-to-text generation in comparison to other systems for both 
\textsc{RotoWire} and MLB datasets.

\section{Discussion}

In this work we presented a plan-and-generate approach for
data-to-text generation which consists of a macro planning stage
representing high-level document organization in terms of structure
and content, followed by a text generation stage.  Extensive automatic
and human evaluation shows that our approach achieves better results
than existing state-of-the-art models and generates summaries which
are factual, coherent, and concise. %
Our results show that macro planning is more advantageous for
generation tasks expected to produce longer texts with multiple
discourse units, and could be easily extended to other sports domains
such as cricket \cite{kelly-etal-2009-investigating} or American
football \cite{barzilay-lapata-2005-collective}.
Other approaches focusing on micro planning
\cite{DBLP:journals/corr/abs-1809-00582,moryossef-etal-2019-step}
might be better tailored for generating shorter texts. There has been
a surge of datasets recently focusing on single-paragraph outputs and
the task of content selection such as E2E
\cite{novikova-etal-2017-e2e}, WebNLG
\cite{gardent-etal-2017-creating}, and WikiBio
\cite{lebret-etal-2016-neural,perez-beltrachini-lapata-2018-bootstrapping}.
We note that in our model content selection takes place during macro
planning \emph{and} text
generation. %
The results in Table~\ref{tbl:results-with-ie-test} show that Macro
achieves the highest CS F-measure on both datasets indicating that the
document as a whole and individual sentences discuss appropriate
content.

Throughout our experiments we observed that template-based systems
score poorly in terms of CS (but also CO and BLEU). This is primarily
due to the inflexibility of the template approach which is limited to
the discussion of a fixed number of (high-scoring) players. Yet, human
writers (and neural models to a certain extent), synthesize summaries
taking into account the particulars of a specific game (where some
players might be more important than others even if they scored less)
and are able to override global defaults.  Template sentences are
fluent on their own, but since it is not possible to perform
aggregation \cite{DBLP:journals/corr/cmp-lg-9504013}, the whole summary appears stilted,
it lacks coherence and variability, contributing to low BLEU scores.
The template baseline is worse for MLB than \textsc{RotoWire} which
reflects the greater difficulty to manually create a good template for
MLB. %
Overall, we observe that neural models are more fluent and coherent,
being able to learn a better ordering of facts which is in turn
reflected in better CO scores. 

Despite promising results, there is ample room to improve macro
planning, especially in terms of the precision of RG (see
Table~\ref{tbl:results-with-ie-test}, P\% column of RG). We should not
underestimate that Macro must handle relatively long inputs (the
average input length in the MLB development set is~\textasciitilde
3100 tokens) which are challenging for the attention
mechanism. Consider the following output of our model on the MLB
dataset: \textsl{Ramirez's two-run double off Joe Blanton tied it in
  the sixth, and Brandon Moss added a two-out RBI single off
  \textcolor{red}{Alan Embree} to give Boston a 3-2 lead}. Here, the
name of the pitcher should have been {\it Joe Blanton} instead of
\textsl{Alan Embree}. In fact, \textsl{Alan Embree} is the pitcher for
the following play in the half inning.
In this case, attention diffuses over the relatively long MLB macro
plan, leading to inaccurate content selection. We could alleviate
this problem by adopting a noisy channel decomposition
\cite{yee-etal-2019-simple,doi:10.1162/tacla00319}, i.e.,~by
learning two different distributions: a conditional model which
provides the probability of {\it translating} a paragraph plan to text
and a language model which provides an unconditional estimate of the
output (i.e.,~the whole game summary). However, we leave this to
future work.

For \textsc{RotoWire}, the main source of errors is the model's
inability to understand numbers. For example, Macro generates the
following output \textsl{The Lakers were the superior shooters in this
  game, going 48 percent from the field and \textcolor{red}{30}
  percent from the three-point line, while the Jazz went 47 percent
  from the floor and 30 percent from beyond the arc.}. Here,
\textsl{30 percent} should have been \textsl{24 percent} for the
Lakers but the language model expects a higher score for the
\textsl{three-point line}, and since \textsl{24} is low (especially
compared to 30 scored by the Jazz), it simply copies \textsl{30}
scored by the Jazz instead. A mechanism for learning better
representations for numbers \cite{wallace-etal-2019-nlp} or executing
operations such as argmax or minus \cite{nie-etal-2018-operation}
should help alleviate this problem.

Finally, although our focus so far has been on learning document plans
from data, the decoupling of planning from generation allows to
flexibly generate output according to specification. For example, we
could feed the model with manually constructed macro plans,
consequently controlling the information content and structure of the
output summary (e.g., for generating short or long texts, or focusing
on specific aspects of the game).

\section*{Acknowledgements}
We thank the Action Editor, Claire Gardent, and the three anonymous reviewers 
for their constructive feedback. 
We also thank Laura Perez-Beltrachini for her comments on an earlier draft of this paper, and
Parag Jain, Hao Zheng, Stefanos Angelidis and Yang Liu for
helpful discussions. 
We acknowledge the financial support of the European Research Council 
(Lapata; award number 681760, ``Translating Multiple Modalities into Text'').

\bibliography{anthology,tacl2020}

\begin{thebibliography}{66}
\expandafter\ifx\csname natexlab\endcsname\relax\def\natexlab#1{#1}\fi

\bibitem[{Bahdanau et~al.(2015)Bahdanau, Cho, and
  Bengio}]{DBLP:journals/corr/BahdanauCB14}
Dzmitry Bahdanau, Kyunghyun Cho, and Yoshua Bengio. 2015.
\newblock \href {http://arxiv.org/abs/1409.0473} {Neural machine translation by
  jointly learning to align and translate}.
\newblock In \emph{3rd International Conference on Learning Representations,
  {ICLR} 2015, San Diego, CA, USA, May 7-9, 2015, Conference Track
  Proceedings}.

\bibitem[{Barzilay and Lapata(2005)}]{barzilay-lapata-2005-collective}
Regina Barzilay and Mirella Lapata. 2005.
\newblock \href {https://www.aclweb.org/anthology/H05-1042} {Collective content
  selection for concept-to-text generation}.
\newblock In \emph{Proceedings of Human Language Technology Conference and
  Conference on Empirical Methods in Natural Language Processing}, pages
  331--338, Vancouver, British Columbia, Canada. Association for Computational
  Linguistics.

\bibitem[{Bird et~al.(2009)Bird, Klein, and Loper}]{BirdKleinLoper09}
Steven Bird, Ewan Klein, and Edward Loper. 2009.
\newblock \emph{{Natural Language Processing with Python}}.
\newblock O'Reilly Media.

\bibitem[{Castro~Ferreira et~al.(2019)Castro~Ferreira, van~der Lee, van
  Miltenburg, and Krahmer}]{castro-ferreira-etal-2019-neural}
Thiago Castro~Ferreira, Chris van~der Lee, Emiel van Miltenburg, and Emiel
  Krahmer. 2019.
\newblock \href {https://doi.org/10.18653/v1/D19-1052} {Neural data-to-text
  generation: A comparison between pipeline and end-to-end architectures}.
\newblock In \emph{Proceedings of the 2019 Conference on Empirical Methods in
  Natural Language Processing and the 9th International Joint Conference on
  Natural Language Processing (EMNLP-IJCNLP)}, pages 552--562, Hong Kong,
  China. Association for Computational Linguistics.

\bibitem[{Chafe(1979)}]{Chafe:1979}
Wallace~L. Chafe. 1979.
\newblock The flow of thought and the flow of language.
\newblock In Talmy Giv\'{o}n, editor, \emph{Syntax and Semantics}, volume~12,
  pages 159--181. Academic Press Inc.

\bibitem[{Chen and Bansal(2018)}]{chen-bansal-2018-fast}
Yen-Chun Chen and Mohit Bansal. 2018.
\newblock \href {https://doi.org/10.18653/v1/P18-1063} {Fast abstractive
  summarization with reinforce-selected sentence rewriting}.
\newblock In \emph{Proceedings of the 56th Annual Meeting of the Association
  for Computational Linguistics (Volume 1: Long Papers)}, pages 675--686,
  Melbourne, Australia. Association for Computational Linguistics.

\bibitem[{Dale(1989)}]{dale1989generating}
Robert Dale. 1989.
\newblock Generating referring expressions in a domain of objects and
  processes.

\bibitem[{Duboue and McKeown(2002)}]{duboue-mckeown-2002-content}
Pablo Duboue and Kathleen McKeown. 2002.
\newblock \href {https://www.aclweb.org/anthology/W02-2112} {Content planner
  construction via evolutionary algorithms and a corpus-based fitness
  function}.
\newblock In \emph{Proceedings of the International Natural Language Generation
  Conference}, pages 89--96, Harriman, New York, USA. Association for
  Computational Linguistics.

\bibitem[{Duboue and McKeown(2001)}]{duboue-mckeown-2001-empirically}
Pablo~A. Duboue and Kathleen~R. McKeown. 2001.
\newblock \href {https://doi.org/10.3115/1073012.1073035} {Empirically
  estimating order constraints for content planning in generation}.
\newblock In \emph{Proceedings of the 39th Annual Meeting of the Association
  for Computational Linguistics}, pages 172--179, Toulouse, France. Association
  for Computational Linguistics.

\bibitem[{Duchi et~al.(2011)Duchi, Hazan, and
  Singer}]{DBLP:journals/jmlr/DuchiHS11}
John~C. Duchi, Elad Hazan, and Yoram Singer. 2011.
\newblock \href {http://dl.acm.org/citation.cfm?id=2021068} {Adaptive
  subgradient methods for online learning and stochastic optimization}.
\newblock \emph{Journal of Machine Learning Research}, 12:2121--2159.

\bibitem[{Fan et~al.(2018)Fan, Grangier, and Auli}]{fan-etal-2018-controllable}
Angela Fan, David Grangier, and Michael Auli. 2018.
\newblock \href {https://doi.org/10.18653/v1/W18-2706} {Controllable
  abstractive summarization}.
\newblock In \emph{Proceedings of the 2nd Workshop on Neural Machine
  Translation and Generation}, pages 45--54, Melbourne, Australia. Association
  for Computational Linguistics.

\bibitem[{Gardent et~al.(2017)Gardent, Shimorina, Narayan, and
  Perez-Beltrachini}]{gardent-etal-2017-creating}
Claire Gardent, Anastasia Shimorina, Shashi Narayan, and Laura
  Perez-Beltrachini. 2017.
\newblock \href {https://doi.org/10.18653/v1/P17-1017} {Creating training
  corpora for {NLG} micro-planners}.
\newblock In \emph{Proceedings of the 55th Annual Meeting of the Association
  for Computational Linguistics (Volume 1: Long Papers)}, pages 179--188,
  Vancouver, Canada. Association for Computational Linguistics.

\bibitem[{Gatt and Krahmer(2018)}]{DBLP:journals/jair/GattK18}
Albert Gatt and Emiel Krahmer. 2018.
\newblock \href {https://doi.org/10.1613/jair.5477} {Survey of the state of the
  art in natural language generation: Core tasks, applications and evaluation}.
\newblock \emph{J. Artif. Intell. Res.}, 61:65--170.

\bibitem[{Gong et~al.(2019)Gong, Feng, Qin, and Liu}]{gong-etal-2019-table}
Heng Gong, Xiaocheng Feng, Bing Qin, and Ting Liu. 2019.
\newblock \href {https://doi.org/10.18653/v1/D19-1310} {Table-to-text
  generation with effective hierarchical encoder on three dimensions (row,
  column and time)}.
\newblock In \emph{Proceedings of the 2019 Conference on Empirical Methods in
  Natural Language Processing and the 9th International Joint Conference on
  Natural Language Processing (EMNLP-IJCNLP)}, pages 3143--3152, Hong Kong,
  China. Association for Computational Linguistics.

\bibitem[{Gu et~al.(2016)Gu, Lu, Li, and Li}]{gu-etal-2016-incorporating}
Jiatao Gu, Zhengdong Lu, Hang Li, and Victor~O.K. Li. 2016.
\newblock \href {https://doi.org/10.18653/v1/P16-1154} {Incorporating copying
  mechanism in sequence-to-sequence learning}.
\newblock In \emph{Proceedings of the 54th Annual Meeting of the Association
  for Computational Linguistics (Volume 1: Long Papers)}, pages 1631--1640,
  Berlin, Germany. Association for Computational Linguistics.

\bibitem[{Gulcehre et~al.(2016)Gulcehre, Ahn, Nallapati, Zhou, and
  Bengio}]{gulcehre-etal-2016-pointing}
Caglar Gulcehre, Sungjin Ahn, Ramesh Nallapati, Bowen Zhou, and Yoshua Bengio.
  2016.
\newblock \href {https://doi.org/10.18653/v1/P16-1014} {Pointing the unknown
  words}.
\newblock In \emph{Proceedings of the 54th Annual Meeting of the Association
  for Computational Linguistics (Volume 1: Long Papers)}, pages 140--149,
  Berlin, Germany. Association for Computational Linguistics.

\bibitem[{Halliday and Hasan(1976)}]{Halliday:Hasan:76}
M.~A.~K. Halliday and Ruqaiya Hasan. 1976.
\newblock \emph{Cohesion in English}.
\newblock Longman, London.

\bibitem[{Hochreiter and Schmidhuber(1997)}]{hochreiter1997long}
Sepp Hochreiter and J\"{u}rgen Schmidhuber. 1997.
\newblock Long {S}hort-{T}erm {M}emory.
\newblock \emph{Neural Computation}, 9:1735--1780.

\bibitem[{Hovy(1993)}]{hovy1993automated}
Eduard~H Hovy. 1993.
\newblock Automated discourse generation using discourse structure relations.
\newblock \emph{Artificial intelligence}, 63(1-2):341--385.

\bibitem[{Iso et~al.(2019)Iso, Uehara, Ishigaki, Noji, Aramaki, Kobayashi,
  Miyao, Okazaki, and Takamura}]{iso-etal-2019-learning}
Hayate Iso, Yui Uehara, Tatsuya Ishigaki, Hiroshi Noji, Eiji Aramaki, Ichiro
  Kobayashi, Yusuke Miyao, Naoaki Okazaki, and Hiroya Takamura. 2019.
\newblock \href {https://doi.org/10.18653/v1/P19-1202} {Learning to select,
  track, and generate for data-to-text}.
\newblock In \emph{Proceedings of the 57th Annual Meeting of the Association
  for Computational Linguistics}, pages 2102--2113, Florence, Italy.
  Association for Computational Linguistics.

\bibitem[{Kan and McKeown(2002)}]{kan-mckeown-2002-corpus}
Min-Yen Kan and Kathleen~R. McKeown. 2002.
\newblock \href {https://www.aclweb.org/anthology/W02-2101} {Corpus-trained
  text generation for summarization}.
\newblock In \emph{Proceedings of the International Natural Language Generation
  Conference}, pages 1--8, Harriman, New York, USA. Association for
  Computational Linguistics.

\bibitem[{Kelly et~al.(2009)Kelly, Copestake, and
  Karamanis}]{kelly-etal-2009-investigating}
Colin Kelly, Ann Copestake, and Nikiforos Karamanis. 2009.
\newblock \href {https://www.aclweb.org/anthology/W09-0623} {Investigating
  content selection for language generation using machine learning}.
\newblock In \emph{Proceedings of the 12th {E}uropean Workshop on Natural
  Language Generation ({ENLG} 2009)}, pages 130--137, Athens, Greece.
  Association for Computational Linguistics.

\bibitem[{Kikuchi et~al.(2016)Kikuchi, Neubig, Sasano, Takamura, and
  Okumura}]{kikuchi-etal-2016-controlling}
Yuta Kikuchi, Graham Neubig, Ryohei Sasano, Hiroya Takamura, and Manabu
  Okumura. 2016.
\newblock \href {https://doi.org/10.18653/v1/D16-1140} {Controlling output
  length in neural encoder-decoders}.
\newblock In \emph{Proceedings of the 2016 Conference on Empirical Methods in
  Natural Language Processing}, pages 1328--1338, Austin, Texas. Association
  for Computational Linguistics.

\bibitem[{Klein et~al.(2017)Klein, Kim, Deng, Senellart, and
  Rush}]{klein-etal-2017-opennmt}
Guillaume Klein, Yoon Kim, Yuntian Deng, Jean Senellart, and Alexander Rush.
  2017.
\newblock \href {https://www.aclweb.org/anthology/P17-4012} {{O}pen{NMT}:
  Open-source toolkit for neural machine translation}.
\newblock In \emph{Proceedings of {ACL} 2017, System Demonstrations}, pages
  67--72, Vancouver, Canada. Association for Computational Linguistics.

\bibitem[{Konstas and Lapata(2013)}]{konstas-lapata-2013-inducing}
Ioannis Konstas and Mirella Lapata. 2013.
\newblock \href {https://www.aclweb.org/anthology/D13-1157} {Inducing document
  plans for concept-to-text generation}.
\newblock In \emph{Proceedings of the 2013 Conference on Empirical Methods in
  Natural Language Processing}, pages 1503--1514, Seattle, Washington, USA.
  Association for Computational Linguistics.

\bibitem[{Kukich(1983)}]{P83-1022}
Karen Kukich. 1983.
\newblock \href
  {http://aclanthology.coli.uni-saarland.de/pdf/P/P83/P83-1022.pdf} {Design of
  a knowledge-based report generator}.
\newblock In \emph{21st Annual Meeting of the Association for Computational
  Linguistics}.

\bibitem[{Laha et~al.(2020)Laha, Jain, Mishra, and
  Sankaranarayanan}]{doi:10.1162/colia00363}
Anirban Laha, Parag Jain, Abhijit Mishra, and Karthik Sankaranarayanan. 2020.
\newblock \href {https://doi.org/10.1162/coli\_a\_00363} {Scalable
  micro-planned generation of discourse from structured data}.
\newblock \emph{Computational Linguistics}, 45(4):737--763.

\bibitem[{Lebret et~al.(2016)Lebret, Grangier, and
  Auli}]{lebret-etal-2016-neural}
R{\'e}mi Lebret, David Grangier, and Michael Auli. 2016.
\newblock \href {https://doi.org/10.18653/v1/D16-1128} {Neural text generation
  from structured data with application to the biography domain}.
\newblock In \emph{Proceedings of the 2016 Conference on Empirical Methods in
  Natural Language Processing}, pages 1203--1213, Austin, Texas. Association
  for Computational Linguistics.

\bibitem[{van~der Lee et~al.(2019)van~der Lee, Gatt, van Miltenburg, Wubben,
  and Krahmer}]{van-der-lee-etal-2019-best}
Chris van~der Lee, Albert Gatt, Emiel van Miltenburg, Sander Wubben, and Emiel
  Krahmer. 2019.
\newblock \href {https://doi.org/10.18653/v1/W19-8643} {Best practices for the
  human evaluation of automatically generated text}.
\newblock In \emph{Proceedings of the 12th International Conference on Natural
  Language Generation}, pages 355--368, Tokyo, Japan. Association for
  Computational Linguistics.

\bibitem[{Longacre(1979)}]{Longacre:1979}
R~E Longacre. 1979.
\newblock The paragraph as a grammatical unit.
\newblock In Talmy Giv\'{o}n, editor, \emph{Syntax and Semantics}, volume~12,
  pages 115--133. Academic Press Inc.

\bibitem[{Louviere et~al.(2015)Louviere, Flynn, and Marley}]{louviere2015best}
Jordan~J. Louviere, Terry~N. Flynn, and A.~A.~J. Marley. 2015.
\newblock \href {https://doi.org/10.1017/CBO9781107337855} {\emph{Best-Worst
  Scaling: Theory, Methods and Applications}}.
\newblock Cambridge University Press.

\bibitem[{Louviere and Woodworth(1991)}]{louviere1991best}
Jordan~J Louviere and George~G Woodworth. 1991.
\newblock Best-worst scaling: A model for the largest difference judgments.
\newblock \emph{University of Alberta: Working Paper}.

\bibitem[{Luong et~al.(2015)Luong, Pham, and
  Manning}]{luong-etal-2015-effective}
Thang Luong, Hieu Pham, and Christopher~D. Manning. 2015.
\newblock \href {https://doi.org/10.18653/v1/D15-1166} {Effective approaches to
  attention-based neural machine translation}.
\newblock In \emph{Proceedings of the 2015 Conference on Empirical Methods in
  Natural Language Processing}, pages 1412--1421, Lisbon, Portugal. Association
  for Computational Linguistics.

\bibitem[{McKeown(1992)}]{mckeown1992text}
Kathleen~R. McKeown. 1992.
\newblock \emph{Text Generation}.
\newblock Studies in Natural Language Processing. Cambridge University Press.

\bibitem[{McKeown et~al.(1997)McKeown, Jordan, Pan, Shaw, and
  Allen}]{mckeown-etal-1997-language}
Kathleen~R. McKeown, Desmond~A. Jordan, Shimei Pan, James Shaw, and Barry~A.
  Allen. 1997.
\newblock \href {https://doi.org/10.3115/974557.974598} {Language generation
  for multimedia healthcare briefings}.
\newblock In \emph{Fifth Conference on Applied Natural Language Processing},
  pages 277--282, Washington, DC, USA. Association for Computational
  Linguistics.

\bibitem[{Mei et~al.(2016)Mei, Bansal, and Walter}]{mei-etal-2016-talk}
Hongyuan Mei, Mohit Bansal, and Matthew~R. Walter. 2016.
\newblock \href {https://doi.org/10.18653/v1/N16-1086} {What to talk about and
  how? selective generation using {LSTM}s with coarse-to-fine alignment}.
\newblock In \emph{Proceedings of the 2016 Conference of the North {A}merican
  Chapter of the Association for Computational Linguistics: Human Language
  Technologies}, pages 720--730, San Diego, California. Association for
  Computational Linguistics.

\bibitem[{Moryossef et~al.(2019)Moryossef, Goldberg, and
  Dagan}]{moryossef-etal-2019-step}
Amit Moryossef, Yoav Goldberg, and Ido Dagan. 2019.
\newblock \href {https://doi.org/10.18653/v1/N19-1236} {{S}tep-by-step:
  {S}eparating planning from realization in neural data-to-text generation}.
\newblock In \emph{Proceedings of the 2019 Conference of the North {A}merican
  Chapter of the Association for Computational Linguistics: Human Language
  Technologies, Volume 1 (Long and Short Papers)}, pages 2267--2277,
  Minneapolis, Minnesota. Association for Computational Linguistics.

\bibitem[{Nie et~al.(2018)Nie, Wang, Yao, Pan, and
  Lin}]{nie-etal-2018-operation}
Feng Nie, Jinpeng Wang, Jin-Ge Yao, Rong Pan, and Chin-Yew Lin. 2018.
\newblock \href {https://doi.org/10.18653/v1/D18-1422} {Operation-guided neural
  networks for high fidelity data-to-text generation}.
\newblock In \emph{Proceedings of the 2018 Conference on Empirical Methods in
  Natural Language Processing}, pages 3879--3889, Brussels, Belgium.
  Association for Computational Linguistics.

\bibitem[{Novikova et~al.(2017)Novikova, Du{\v{s}}ek, and
  Rieser}]{novikova-etal-2017-e2e}
Jekaterina Novikova, Ond{\v{r}}ej Du{\v{s}}ek, and Verena Rieser. 2017.
\newblock \href {https://doi.org/10.18653/v1/W17-5525} {The {E}2{E} dataset:
  New challenges for end-to-end generation}.
\newblock In \emph{Proceedings of the 18th Annual {SIG}dial Meeting on
  Discourse and Dialogue}, pages 201--206, Saarbr{\"u}cken, Germany.
  Association for Computational Linguistics.

\bibitem[{Orme(2009)}]{orme2009maxdiff}
Bryan Orme. 2009.
\newblock Maxdiff analysis: Simple counting, individual-level logit, and hb.
\newblock \emph{Sawtooth Software}.

\bibitem[{Papineni et~al.(2002)Papineni, Roukos, Ward, and
  Zhu}]{papineni-etal-2002-bleu}
Kishore Papineni, Salim Roukos, Todd Ward, and Wei-Jing Zhu. 2002.
\newblock \href {https://doi.org/10.3115/1073083.1073135} {{B}leu: a method for
  automatic evaluation of machine translation}.
\newblock In \emph{Proceedings of the 40th Annual Meeting of the Association
  for Computational Linguistics}, pages 311--318, Philadelphia, Pennsylvania,
  USA. Association for Computational Linguistics.

\bibitem[{Paulus et~al.(2018)Paulus, Xiong, and Socher}]{paulus2018a}
Romain Paulus, Caiming Xiong, and Richard Socher. 2018.
\newblock \href {https://openreview.net/forum?id=HkAClQgA-} {A deep reinforced
  model for abstractive summarization}.
\newblock In \emph{International Conference on Learning Representations}.

\bibitem[{Perez-Beltrachini and
  Lapata(2018)}]{perez-beltrachini-lapata-2018-bootstrapping}
Laura Perez-Beltrachini and Mirella Lapata. 2018.
\newblock \href {https://doi.org/10.18653/v1/N18-1137} {Bootstrapping
  generators from noisy data}.
\newblock In \emph{Proceedings of the 2018 Conference of the North {A}merican
  Chapter of the Association for Computational Linguistics: Human Language
  Technologies, Volume 1 (Long Papers)}, pages 1516--1527, New Orleans,
  Louisiana. Association for Computational Linguistics.

\bibitem[{Puduppully et~al.(2019{\natexlab{a}})Puduppully, Dong, and
  Lapata}]{DBLP:journals/corr/abs-1809-00582}
Ratish Puduppully, Li~Dong, and Mirella Lapata. 2019{\natexlab{a}}.
\newblock \href {https://doi.org/10.1609/aaai.v33i01.33016908} {Data-to-text
  generation with content selection and planning}.
\newblock In \emph{Proceedings of the 33rd AAAI Conference on Artificial
  Intelligence}, Honolulu, Hawaii.

\bibitem[{Puduppully et~al.(2019{\natexlab{b}})Puduppully, Dong, and
  Lapata}]{puduppully-etal-2019-data}
Ratish Puduppully, Li~Dong, and Mirella Lapata. 2019{\natexlab{b}}.
\newblock \href {https://doi.org/10.18653/v1/P19-1195} {Data-to-text generation
  with entity modeling}.
\newblock In \emph{Proceedings of the 57th Annual Meeting of the Association
  for Computational Linguistics}, pages 2023--2035, Florence, Italy.
  Association for Computational Linguistics.

\bibitem[{Radford et~al.(2019)Radford, Wu, Child, Luan, Amodei, and
  Sutskever}]{Radford2019LanguageMA}
Alec Radford, Jeffrey Wu, Rewon Child, David Luan, Dario Amodei, and Ilya
  Sutskever. 2019.
\newblock Language models are unsupervised multitask learners.
\newblock \emph{OpenAI blog}, 1(8):9.

\bibitem[{Rebuffel et~al.(2020)Rebuffel, Soulier, Scoutheeten, and
  Gallinari}]{rebuffel2020hierarchical}
Cl{\'e}ment Rebuffel, Laure Soulier, Geoffrey Scoutheeten, and Patrick
  Gallinari. 2020.
\newblock \href
  {https://link.springer.com/content/pdf/10.1007%2F978-3-030-45439-5_5.pdf} {A
  hierarchical model for data-to-text generation}.
\newblock In \emph{European Conference on Information Retrieval}, pages 65--80.
  Springer.

\bibitem[{Reiter(1995)}]{DBLP:journals/corr/cmp-lg-9504013}
Ehud Reiter. 1995.
\newblock \href {http://arxiv.org/abs/cmp-lg/9504013v1} {{NLG} vs. templates}.
\newblock \emph{CoRR}, cmp-lg/9504013v1.

\bibitem[{Reiter and Dale(1997)}]{Reiter:1997:BAN:974487.974490}
Ehud Reiter and Robert Dale. 1997.
\newblock \href {https://doi.org/10.1017/S1351324997001502} {Building applied
  natural language generation systems}.
\newblock \emph{Nat. Lang. Eng.}, 3(1):57--87.

\bibitem[{Reiter and Dale(2000)}]{reiter-dale:00}
Ehud Reiter and Robert Dale. 2000.
\newblock \href {https://doi.org/10.1017/CBO9780511519857} {\emph{Building
  Natural Language Generation Systems}}.
\newblock Studies in Natural Language Processing. Cambridge University Press.

\bibitem[{Saleh et~al.(2019)Saleh, Berard, Calapodescu, and
  Besacier}]{saleh-etal-2019-naver}
Fahimeh Saleh, Alexandre Berard, Ioan Calapodescu, and Laurent Besacier. 2019.
\newblock \href {https://doi.org/10.18653/v1/D19-5631} {Naver labs {E}urope{'}s
  systems for the document-level generation and translation task at {WNGT}
  2019}.
\newblock In \emph{Proceedings of the 3rd Workshop on Neural Generation and
  Translation}, pages 273--279, Hong Kong. Association for Computational
  Linguistics.

\bibitem[{Sennrich et~al.(2016)Sennrich, Haddow, and
  Birch}]{sennrich-etal-2016-neural}
Rico Sennrich, Barry Haddow, and Alexandra Birch. 2016.
\newblock \href {https://doi.org/10.18653/v1/P16-1162} {Neural machine
  translation of rare words with subword units}.
\newblock In \emph{Proceedings of the 54th Annual Meeting of the Association
  for Computational Linguistics (Volume 1: Long Papers)}, pages 1715--1725,
  Berlin, Germany. Association for Computational Linguistics.

\bibitem[{Shao et~al.(2019)Shao, Huang, Wen, Xu, and Zhu}]{shao-etal-2019-long}
Zhihong Shao, Minlie Huang, Jiangtao Wen, Wenfei Xu, and Xiaoyan Zhu. 2019.
\newblock \href {https://doi.org/10.18653/v1/D19-1321} {Long and diverse text
  generation with planning-based hierarchical variational model}.
\newblock In \emph{Proceedings of the 2019 Conference on Empirical Methods in
  Natural Language Processing and the 9th International Joint Conference on
  Natural Language Processing (EMNLP-IJCNLP)}, pages 3257--3268, Hong Kong,
  China. Association for Computational Linguistics.

\bibitem[{Sutskever et~al.(2014)Sutskever, Vinyals, and
  Le}]{sutskever2014sequence}
Ilya Sutskever, Oriol Vinyals, and Quoc~V Le. 2014.
\newblock \href
  {https://proceedings.neurips.cc/paper/2014/file/a14ac55a4f27472c5d894ec1c3c743d2-Paper.pdf}
  {Sequence to sequence learning with neural networks}.
\newblock In \emph{Advances in Neural Information Processing Systems},
  volume~27, pages 3104--3112. Curran Associates, Inc.

\bibitem[{Takeno et~al.(2017)Takeno, Nagata, and
  Yamamoto}]{takeno-etal-2017-controlling}
Shunsuke Takeno, Masaaki Nagata, and Kazuhide Yamamoto. 2017.
\newblock \href {https://www.aclweb.org/anthology/W17-5702} {Controlling target
  features in neural machine translation via prefix constraints}.
\newblock In \emph{Proceedings of the 4th Workshop on {A}sian Translation
  ({WAT}2017)}, pages 55--63, Taipei, Taiwan. Asian Federation of Natural
  Language Processing.

\bibitem[{Tian et~al.(2019)Tian, Narayan, Sellam, and
  Parikh}]{DBLP:journals/corr/abs-1910-08684}
Ran Tian, Shashi Narayan, Thibault Sellam, and Ankur~P. Parikh. 2019.
\newblock \href {http://arxiv.org/abs/1910.08684v2} {Sticking to the facts:
  Confident decoding for faithful data-to-text generation}.
\newblock \emph{CoRR}, abs/1910.08684v2.

\bibitem[{Vaswani et~al.(2017)Vaswani, Shazeer, Parmar, Uszkoreit, Jones,
  Gomez, Kaiser, and Polosukhin}]{NIPS2017_7181}
Ashish Vaswani, Noam Shazeer, Niki Parmar, Jakob Uszkoreit, Llion Jones,
  Aidan~N Gomez, \L~ukasz Kaiser, and Illia Polosukhin. 2017.
\newblock \href
  {http://papers.nips.cc/paper/7181-attention-is-all-you-need.pdf} {Attention
  is all you need}.
\newblock In I.~Guyon, U.~V. Luxburg, S.~Bengio, H.~Wallach, R.~Fergus,
  S.~Vishwanathan, and R.~Garnett, editors, \emph{Advances in Neural
  Information Processing Systems 30}, pages 5998--6008. Curran Associates, Inc.

\bibitem[{Vinyals et~al.(2015)Vinyals, Fortunato, and Jaitly}]{NIPS2015_5866}
Oriol Vinyals, Meire Fortunato, and Navdeep Jaitly. 2015.
\newblock \href {http://papers.nips.cc/paper/5866-pointer-networks.pdf}
  {Pointer networks}.
\newblock In C.~Cortes, N.~D. Lawrence, D.~D. Lee, M.~Sugiyama, and R.~Garnett,
  editors, \emph{Advances in Neural Information Processing Systems 28}, pages
  2692--2700. Curran Associates, Inc.

\bibitem[{Wallace et~al.(2019)Wallace, Wang, Li, Singh, and
  Gardner}]{wallace-etal-2019-nlp}
Eric Wallace, Yizhong Wang, Sujian Li, Sameer Singh, and Matt Gardner. 2019.
\newblock \href {https://doi.org/10.18653/v1/D19-1534} {Do {NLP} models know
  numbers? probing numeracy in embeddings}.
\newblock In \emph{Proceedings of the 2019 Conference on Empirical Methods in
  Natural Language Processing and the 9th International Joint Conference on
  Natural Language Processing (EMNLP-IJCNLP)}, pages 5307--5315, Hong Kong,
  China. Association for Computational Linguistics.

\bibitem[{Williams and Peng(1990)}]{DBLP:journals/neco/WilliamsP90}
Ronald~J. Williams and Jing Peng. 1990.
\newblock \href {https://doi.org/10.1162/neco.1990.2.4.490} {An efficient
  gradient-based algorithm for on-line training of recurrent network
  trajectories}.
\newblock \emph{Neural Computation}, 2(4):490--501.

\bibitem[{Wiseman et~al.(2017)Wiseman, Shieber, and
  Rush}]{wiseman-etal-2017-challenges}
Sam Wiseman, Stuart Shieber, and Alexander Rush. 2017.
\newblock \href {https://doi.org/10.18653/v1/D17-1239} {Challenges in
  data-to-document generation}.
\newblock In \emph{Proceedings of the 2017 Conference on Empirical Methods in
  Natural Language Processing}, pages 2253--2263, Copenhagen, Denmark.
  Association for Computational Linguistics.

\bibitem[{Wu et~al.(2016)Wu, Schuster, Chen, Le, Norouzi, Macherey, Krikun,
  Cao, Gao, Macherey, Klingner, Shah, Johnson, Liu, Kaiser, Gouws, Kato, Kudo,
  Kazawa, Stevens, Kurian, Patil, Wang, Young, Smith, Riesa, Rudnick, Vinyals,
  Corrado, Hughes, and Dean}]{DBLP:journals/corr/WuSCLNMKCGMKSJL16}
Yonghui Wu, Mike Schuster, Zhifeng Chen, Quoc~V. Le, Mohammad Norouzi, Wolfgang
  Macherey, Maxim Krikun, Yuan Cao, Qin Gao, Klaus Macherey, Jeff Klingner,
  Apurva Shah, Melvin Johnson, Xiaobing Liu, Lukasz Kaiser, Stephan Gouws,
  Yoshikiyo Kato, Taku Kudo, Hideto Kazawa, Keith Stevens, George Kurian,
  Nishant Patil, Wei Wang, Cliff Young, Jason Smith, Jason Riesa, Alex Rudnick,
  Oriol Vinyals, Greg Corrado, Macduff Hughes, and Jeffrey Dean. 2016.
\newblock \href {http://arxiv.org/abs/1609.08144} {Google's neural machine
  translation system: Bridging the gap between human and machine translation}.
\newblock \emph{CoRR}, abs/1609.08144v2.

\bibitem[{Yang et~al.(2016)Yang, Yang, Dyer, He, Smola, and
  Hovy}]{yang-etal-2016-hierarchical}
Zichao Yang, Diyi Yang, Chris Dyer, Xiaodong He, Alex Smola, and Eduard Hovy.
  2016.
\newblock \href {https://doi.org/10.18653/v1/N16-1174} {Hierarchical attention
  networks for document classification}.
\newblock In \emph{Proceedings of the 2016 Conference of the North {A}merican
  Chapter of the Association for Computational Linguistics: Human Language
  Technologies}, pages 1480--1489, San Diego, California. Association for
  Computational Linguistics.

\bibitem[{Yee et~al.(2019)Yee, Dauphin, and Auli}]{yee-etal-2019-simple}
Kyra Yee, Yann Dauphin, and Michael Auli. 2019.
\newblock \href {https://doi.org/10.18653/v1/D19-1571} {Simple and effective
  noisy channel modeling for neural machine translation}.
\newblock In \emph{Proceedings of the 2019 Conference on Empirical Methods in
  Natural Language Processing and the 9th International Joint Conference on
  Natural Language Processing (EMNLP-IJCNLP)}, pages 5696--5701, Hong Kong,
  China. Association for Computational Linguistics.

\bibitem[{Yu et~al.(2020)Yu, Sartran, Stokowiec, Ling, Kong, Blunsom, and
  Dyer}]{doi:10.1162/tacla00319}
Lei Yu, Laurent Sartran, Wojciech Stokowiec, Wang Ling, Lingpeng Kong, Phil
  Blunsom, and Chris Dyer. 2020.
\newblock \href {https://doi.org/10.1162/tacl\_a\_00319} {Better document-level
  machine translation with bayes' rule}.
\newblock \emph{Transactions of the Association for Computational Linguistics},
  8:346--360.

\bibitem[{Zadrozny and Jensen(1991)}]{zadrozny-jensen-1991-semantics}
Wlodek Zadrozny and Karen Jensen. 1991.
\newblock \href {https://www.aclweb.org/anthology/J91-2003} {Semantics of
  paragraphs}.
\newblock \emph{Computational Linguistics}, 17(2):171--210.

\end{thebibliography}
\bibliographystyle{acl_natbib}

\end{document}